\documentclass[lettersize,journal]{IEEEtran}
\usepackage{amsmath,amsfonts}
\usepackage{algorithmic}
\usepackage{algorithm}
\usepackage{array}
\usepackage{hyperref}       
\usepackage[caption=false,font=normalsize,labelfont=sf,textfont=sf]{subfig}
\usepackage{textcomp}
\usepackage{stfloats}
\usepackage{url}
\usepackage{verbatim}
\usepackage{graphicx}
\usepackage{cite}
\usepackage{tabularx}

\newcolumntype{Y}{>{\centering\arraybackslash}X}
\newcolumntype{C}{>{\centering\arraybackslash}X}
\newcolumntype{M}{>{\centering\arraybackslash}p{1.35cm}}
\usepackage{booktabs}
\usepackage{mathrsfs}
\usepackage{multirow}
\usepackage{makecell}
\usepackage[table]{xcolor}     
\usepackage[most]{tcolorbox}

\usepackage{amsthm}       
\newcommand{\Embf}[1]{\textcolor{red}{\textbf{#1}}}

\newcommand{\cellsecond}[1]{\cellcolor{orange!40}#1}
\newcommand{\cellbest}[1]{\cellcolor{red!40}#1}
\newtcolorbox{blackclaim}[1][]{%
  colback=white,
  colframe=black,
  boxrule=0.8pt,
  sharp corners,
  left=6pt,right=6pt,top=4pt,bottom=4pt,
  #1
}

\theoremstyle{plain}
\newtheorem{theorem}{Theorem}[section]
\newtheorem{proposition}[theorem]{Proposition}

\theoremstyle{definition}
\newtheorem{definition}[theorem]{Definition}

\theoremstyle{remark}

\begin{document}
\bstctlcite{IEEEexample:BSTcontrol}
\newcommand{\Paragraph}[1]{\vspace{2mm} \noindent \textbf{\textit{#1}}}

\title{SoLAR: Error-Resilient Streamable Long-Horizon Free-Viewpoint Video Reconstruction with Anchor Activation and Latent Recalibration}

\author{Haotian Zhang,~\IEEEmembership{Graduate Student Member,~IEEE}, Xu Mo, Yixin Yu, Guanhua Zhu, Jian Xue,\\
Tongda Xu, Yan Wang, Jiaqi Zhang, Siwei Ma,~\IEEEmembership{Fellow,~IEEE}, and Wen Gao,~\IEEEmembership{Fellow,~IEEE}%
\thanks{H. Zhang, J. Zhang, S. Ma, and W. Gao are with the National Engineering Laboratory for Video Technology, School of Computer Science, Peking University, Beijing 100871, China (e-mail: htzhang25@stu.pku.edu.cn; \{jqzhang, swma, wgao\}@pku.edu.cn).}%
\thanks{X. Mo, Y. Yu,  G. Zhu, and J. Xue are with the School of Computer Science, Jilin University, Changchun 130012, China (e-mail: \{moxu2123, yuyx1723, zhugh2424, xuejian2124\}@mails.jlu.edu.cn).}%
\thanks{T. Xu and Y. Wang are with Institute for AI Industry Research (AIR), Tsinghua University, Beijing 100084, China (e-mail: x.tongda@nyu.edu; wangyan@air.tsinghua.edu.cn).}%
}

\markboth{IEEE Transactions on Pattern Analysis and Machine Intelligence,~Vol.~XX, No.~XX, XXXX~2026}%
{Shell \MakeLowercase{\textit{et al.}}:SoLAR: Error-Resilient Streamable Long-Horizon Free-Viewpoint Video Reconstruction with Anchor Activation and Latent Recalibration}

\maketitle

\begin{figure*}[t]
  \centering
  \begin{minipage}[t]{0.56\textwidth}
    \centering
    \includegraphics[
      width=\linewidth,
      height=4.5cm,
      keepaspectratio,
      trim=0 0 0 0,
      clip
    ]{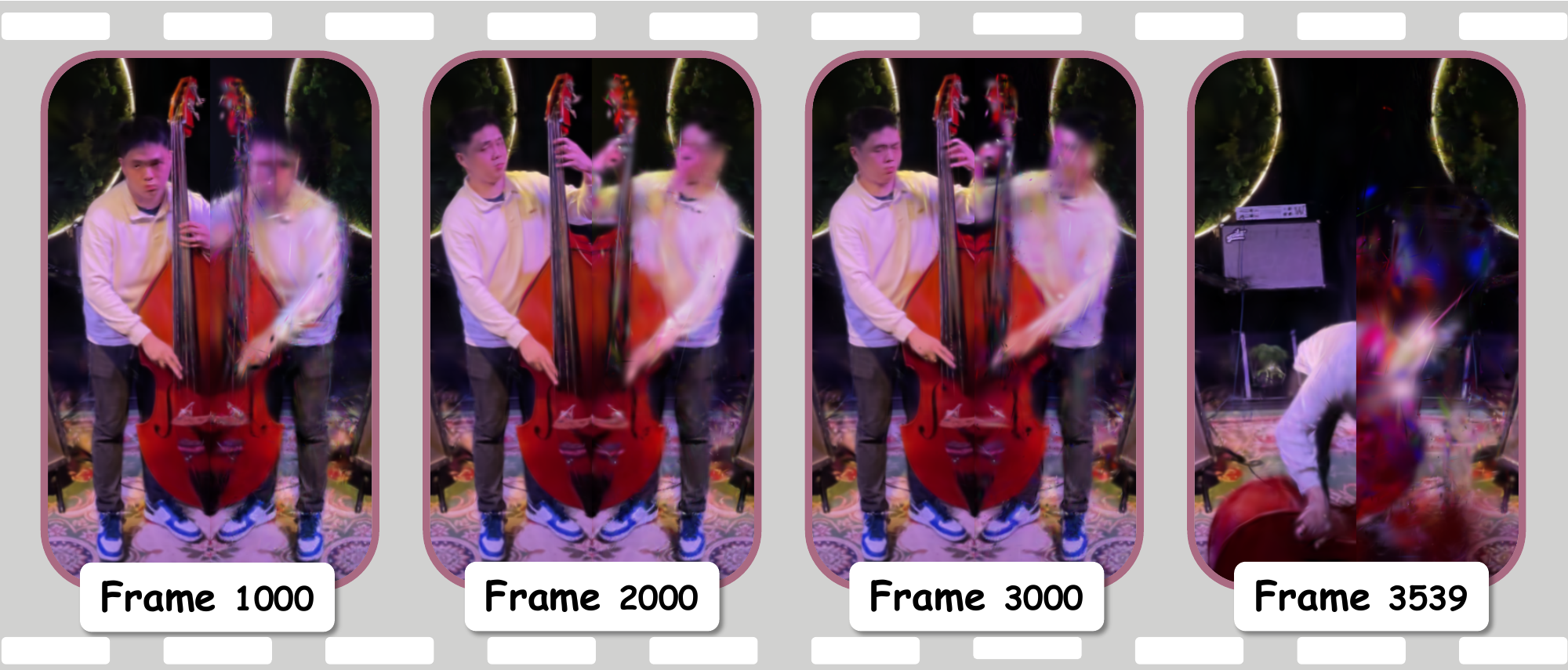}
  \end{minipage}\hfill
  \begin{minipage}[t]{0.40\textwidth}
    \centering
    \includegraphics[
      width=\linewidth,
      height=4.5cm,
      keepaspectratio,
      trim=0 0 0 0,
      clip
    ]{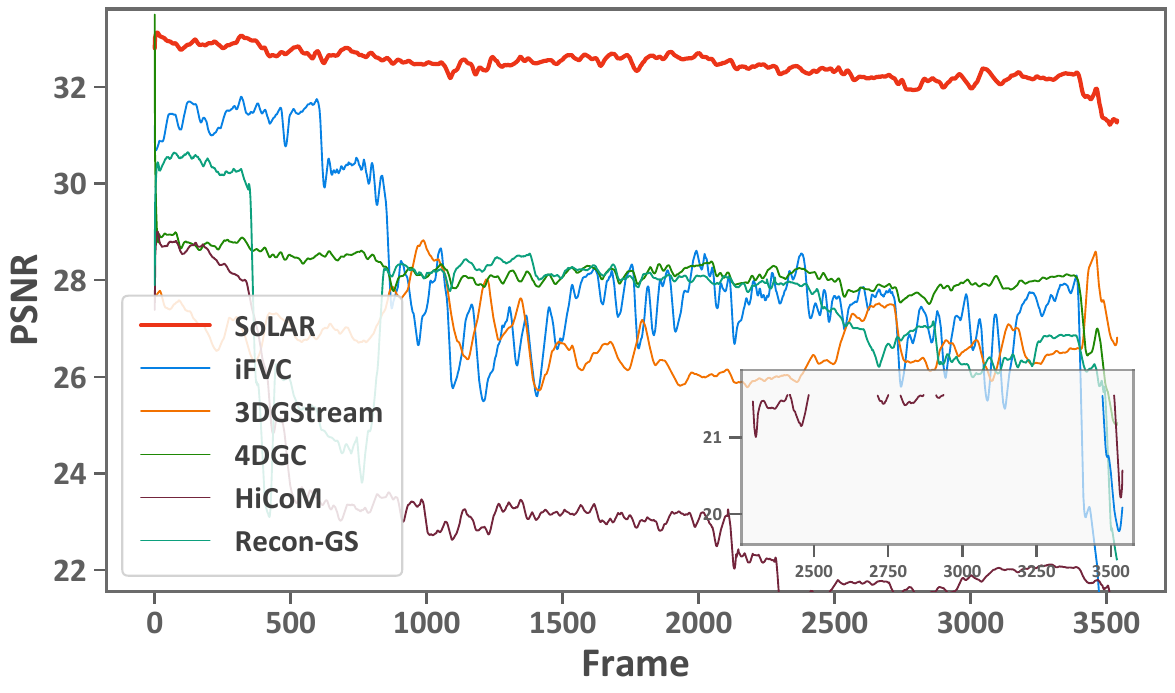}
  \end{minipage}
\vspace{-3mm}
  \caption{
    \textbf{(a) Qualitative results of a long free-viewpoint video (FVV).} Split-view frames compare \textbf{SoLAR} (left) with iFVC (right): iFVC shows visible visual deterioration over time, while \textbf{SoLAR} preserves fine details and high-fidelity rendering. \textbf{(b) Reconstruction quality trends.} Competing methods suffer severe performance decay from error propagation, whereas \textbf{SoLAR} mitigates this, showing robust \textbf{error-resilient} performance in long-horizon videos.
  }
  \label{fig:teaser}
\end{figure*}
\begin{abstract}
Free-Viewpoint Video (FVV) has emerged as a cornerstone of next-generation immersive media systems and attracted widespread attention.
Previous methods primarily focus on short video sequences and suffer from significant performance degradation when processing long-horizon free-viewpoint video (LFVV).
Motivated by bit allocation theory, we analyze dynamic-anchor-based volumetric video representation within a rate-distortion optimization framework and propose \textbf{SoLAR}, which is the first error-resilient streamable FVV framework that maintains stable reconstruction quality on long sequences without requiring group-of-pictures partitioning.
We propose the Anchor Activation Dynamics (AAD), which enables dynamic anchors to model non-rigid transformations by dynamically activating informative anchors and suppressing redundant ones.
Furthermore, we introduce Latent Discrepancy Aware Recalibration (LaDAR), which is a mechanism to identify discrepancies between latent representations and recalibrate the correspondences encoded in the network, effectively mitigating error propagation in LFVV without compromising real-time performance or storage compactness.
Extensive experiments demonstrate that \textbf{SoLAR} achieves state-of-the-art reconstruction performance while maintaining minimum storage overhead, which  provides a new direction for LFVV reconstruction and advances the practical deployment of immersive systems. Demo free-viewpoint videos are provided in the supplementary material.
\end{abstract}


\section{Introduction}

Free-Viewpoint Video (FVV), also referred to as volumetric video, has emerged as a cornerstone of next-generation immersive media systems~\cite{zhu2025dynamic,n3dv,streamrf}. 
While this technology enables dynamic scene exploration from arbitrary viewpoints, its pronounced spatio-temporal complexity poses persistent challenges in simultaneously achieving \textit{high reconstruction quality} and \textit{efficient storage}~\cite{zheng20254dgcpro}.
Although offline approaches~\cite{rotor4dgs, spacetimegaussian, xu20244k4d} are capable of achieving high-quality reconstruction, their requirement for access to the complete video sequence during training prevents them from meeting the low-latency demands of real-time applications such as streaming media and immersive live broadcasting. 
To address this limitation, recent methods~\cite{sun20243dgstream,tang2025compressing,gao2024hicom,furecon} have adopted streamable (i.e., on-the-fly) reconstruction frameworks that progressively model inter-frame deformation for frame-by-frame reconstruction, achieving excellent real-time performance. Among these approaches, iFVC~\cite{tang2025compressing} introduces a dynamic-anchor-based volumetric video representation that models the first frame with anchors and represents each subsequent frame with a compact binary transformation cache (BTC), enabling compact storage and high-quality reconstruction. 

Despite these advancements, existing streaming reconstruction models have been primarily evaluated on short video sequences of about 300 frames~\cite{sun20243dgstream,tang2025compressing,gao2024hicom,furecon}. In contrast, the problem of long-horizon free-viewpoint video (LFVV) reconstruction, which often \textit{involves more than \textbf{1,000} frames}, remains largely \textit{underexplored}. However, real-world videos are typically much longer, and this gap in the literature hinders the practical deployment of immersive systems. Therefore, systematic investigation of LFVV reconstruction is essential.
LFVV reconstruction is particularly challenging because objects in the video may undergo \textit{large-amplitude motion} or even \textit{disappear} over time, which imposes significant demands on the \textit{modeling capacity} and \textit{temporal robustness} of reconstruction methods~\cite{xu2024representing}.
Our investigation reveals that critical bottlenecks remain in achieving stable streamable LFVV reconstruction. As shown in Fig.~\ref{fig:teaser}, current state-of-the-art (SOTA) streaming methods suffer from significant performance degradation when processing LFVV. This degradation is primarily caused by the limited \textit{representational capacity} of the model and the \textit{error propagation} through continuous inter-frame transformations. A straightforward approach to address this issue is to partition the long video sequence into several groups of pictures (GOP)~\cite{shaw2024swings}. However, this strategy introduces substantial latency and storage overhead, degrading both the real-time efficiency and overall rate–distortion (RD) performance~\cite{li2016lambda}. Thus, effectively mitigating error propagation in LFVV while preserving compact storage and real-time processing remains a highly \textbf{challenging} and \textbf{underexplored} problem~\cite{zhu2025dynamic}. 

To address the aforementioned challenges, we propose \textbf{SoLAR}, the first error-resilient streamable FVV framework specifically designed for LFVV reconstruction. Unlike existing approaches that often suffer from severe quality degradation as the sequence length increases, \textbf{SoLAR} maintains stable reconstruction quality over long sequences without requiring GOP partitioning, thereby preserving the low-latency and storage-efficient characteristics that are essential for practical streaming scenarios. 
To overcome the limited modeling capacity of existing streaming methods, we propose \textbf{Anchor Activation Dynamics (AAD)}. The key idea is to endow dynamic anchors with the ability to adaptively allocate representational capacity according to evolving scene content, so that informative anchors are activated while redundant ones are suppressed. As a result, AAD enables more effective modeling of non-rigid transformations and complex long-term motion, while simultaneously improving both scene compactness and representational expressiveness.
Furthermore, to mitigate error propagation while preserving compact storage and real-time processing, we draw inspiration from bit allocation theory in traditional video coding~\cite{hu2011rate,li2012rate,li2014lambda,li2016lambda} and analyze dynamic-anchor-based volumetric video representation within a RD optimization framework under rate control. Based on this analysis, we introduce \textbf{Latent Discrepancy Aware Recalibration (LaDAR)}, which identifies discrepancies between latent representations and uses them to recalibrate the correspondences encoded in the network. This design effectively suppresses the accumulation of temporal errors during streaming reconstruction, thereby improving long-horizon stability without sacrificing real-time performance or storage compactness.

Extensive experiments demonstrate that \textbf{SoLAR} achieves SOTA reconstruction quality while maintaining minimum storage overhead. In particular, compared with existing state-of-the-art methods such as Recon-GS~\cite{furecon}, \textbf{SoLAR} improves reconstruction fidelity by approximately \Embf{3}~dB on short sequences, while simultaneously reducing storage overhead by \Embf{6}$\times$. More importantly, the superiority of \textbf{SoLAR} becomes even more evident in the long-horizon setting, where accumulated errors pose substantially greater challenges: on such sequences, the reconstruction gain further increases to approximately \Embf{4}~dB, accompanied by a substantial \Embf{10}$\times$ reduction in storage requirements.
These results indicate that the proposed framework not only improves the conventional quality--efficiency trade-off, but also delivers markedly stronger temporal robustness as the sequence length increases.
By effectively mitigating error propagation without relying on GOP partitioning, \textbf{SoLAR} establishes a practical and scalable solution for LFVV reconstruction, opening a promising direction for long-horizon streamable volumetric video modeling and advancing the real-world deployment of immersive media systems. 

Our main contributions are summarized as follows:
\begin{itemize}
\item We address the previously underexplored problem of long-horizon free-viewpoint video (LFVV) reconstruction and propose \textbf{SoLAR}, the first error-resilient streamable FVV framework designed to maintain stable reconstruction quality over long sequences without resorting to GOP partitioning. By eliminating repeated restarts of temporal modeling, \textbf{SoLAR} preserves the low-latency and storage-efficient properties required by practical streaming and immersive media systems, thereby establishing a new paradigm for streamable LFVV reconstruction.
\item We propose \textbf{Latent Discrepancy Aware Recalibration (LaDAR)}, a principled mechanism for suppressing long-term error accumulation in streaming reconstruction. Specifically, dynamic-anchor-based volumetric video representation is analyzed from the perspective of rate-distortion optimization under rate control, which provides the basis for identifying latent discrepancies that are strongly correlated with temporal degradation. Building on this analysis, LaDAR recalibrates the learned correspondences encoded in the network, thereby effectively mitigating error propagation while preserving compact storage and real-time processing capability.
\item We introduce \textbf{Anchor Activation Dynamics (AAD)} to enhance the modeling capacity of streamable reconstruction for complex long-horizon motions. Instead of relying on a fixed set of active anchors throughout the sequence, AAD enables dynamic anchors to adaptively activate informative components and suppress redundant ones according to evolving scene content. This design improves the ability of the representation to capture non-rigid transformations and large-amplitude motion, while simultaneously promoting compactness and more efficient use of representational capacity.

\item We implement a free-viewpoint rendering pipeline tailored to dynamic-anchor-based volumetric video representation. The pipeline enables reconstructed results to be observed from arbitrary viewpoints, thereby supporting immersive visualization. Demo videos are provided in the supplementary material.

\item We conduct extensive experiments to verify the effectiveness of \textbf{SoLAR} in both reconstruction quality and storage efficiency. The results consistently demonstrate that \textbf{SoLAR} outperforms existing state-of-the-art methods on both short and long sequences, with particularly pronounced advantages in the long-horizon setting where temporal drift and accumulated errors become severe. These findings confirm that the proposed framework provides a stable, efficient, and practically deployable solution for LFVV reconstruction. 
\end{itemize}

The remainder of this paper is organized as follows. Sec.~\ref{sec::related_work} reviews the most relevant prior work. Sec.~\ref{sec:Preliminaries} introduces the preliminaries of the proposed framework. Sec.~\ref{sec:Methodology} presents the proposed SoLAR in detail. Sec.~\ref{section:Experiments} describes the experimental setup and reports the corresponding results and analyses. Finally, Sec.~\ref{sec:conclusion} concludes the paper.
\section{Related Work}
\label{sec::related_work}
\subsection{Static Scene Reconstruction}
Recently, 3D Gaussian Splatting (3DGS)~\cite{3DGS,ren2025octree} has gradually supplanted Neural Radiance Fields (NeRF)-based methods~\cite{NERF} as the dominant approach for static scene reconstruction. Nevertheless, the substantial storage overhead of 3DGS remains a major obstacle to its practical deployment, especially when a large number of Gaussian primitives is required to preserve fine-grained appearance and geometry details~\cite{gao2025deep,liu2026next,wang2022sparse,chen2025hacpp}. This challenge has consequently motivated a growing body of work on more compact 3DGS representations while preserving rendering fidelity~\cite{lee2025compression,fang2025efficient,xiao2025mcgs,niedermayr2024compressed}.

Existing compact 3DGS methods can be broadly categorized into several directions. Some approaches~\cite{lee2024compact,lightgaussian} aim to prune Gaussian primitives, for example by removing uninformative Gaussians using trainable mask mechanisms or gradient-based thresholding strategies. These methods reduce memory and storage consumption by directly suppressing redundant splats, but their compression performance is often closely tied to the quality of the pruning criterion and may sacrifice local details when aggressive pruning is applied. Other works, such as Scaffold-GS~\cite{lu2024scaffold}, achieve effective storage reduction by clustering Gaussians into anchors and implicitly representing Gaussian attributes within each cluster using Multilayer Perceptron (MLP), where Gaussian offsets can be predicted from anchor features to further improve representation compactness. This anchor-based paradigm is attractive because it replaces explicit storage of dense Gaussian attributes with a more structured latent representation. Moreover, HAC~\cite{chen2024hac,chen2025hacpp} further compresses 3DGS by incorporating spatial structure priors into entropy encoders and related compression techniques, demonstrating the importance of exploiting structural redundancy for compact representation.

\subsection{Dynamic Scene Reconstruction}
Dynamic 3D scene reconstruction is one of the most challenging problems in computer vision and can be broadly divided into two paradigms~\cite{zhu2025dynamic,pumarola2021d}. The first paradigm consists of offline methods~\cite{4dgs,wang2025freetimegs,wu20244d,lin2024gaussian,huang2024sc,kwak2025modec}, which train high-quality and temporally consistent dynamic scene representations by accessing the complete video sequence during offline optimization. Since all frames are jointly available during training, such methods can better enforce temporal coherence and often achieve strong reconstruction quality~\cite{li2021neural}. However, the reliance on full-sequence data during training prevents these methods from satisfying the low-latency requirements of real-time applications and makes them less suitable for streamable free-viewpoint video systems~\cite{tang2025compressing,sun20243dgstream}.
The second paradigm adopts streaming or online frameworks that first reconstruct a high-quality initial representation for the first frame and then model subsequent frames through deformation fields~\cite{sun20243dgstream,deformabl3dgs,yan2025instant,wang2023neural}. Compared with offline methods, this setting is more practical for real-time free-viewpoint video transmission because it avoids repeatedly optimizing a complete sequence from scratch. A major challenge in this setting is the substantial storage overhead introduced by per-frame parameters~\cite{zhang2025mega}. To reduce this burden, iFVC~\cite{tang2025compressing} proposes a dynamic-anchor-based volumetric video representation that models the first frame using anchors and represents each subsequent frame with a compact binary transformation cache, thereby significantly reducing storage overhead.
Despite its compact design, iFVC also reveals the core difficulty of streamable dynamic reconstruction. Specifically, the fixed anchor set constrains its ability to represent non-rigid deformations, and the shared Gaussian attribute network may gradually accumulate large discrepancies as the sequence evolves. These limitations lead to insufficient representational capacity and severe error propagation over time. Therefore, although recent streaming methods have improved efficiency, maintaining both compact storage and stable reconstruction quality over long sequences remains an open challenge~\cite{zhu2025dynamic}.

\subsection{Long-Horizon Free-Viewpoint Video Reconstruction}
Despite substantial progress in dynamic scene reconstruction~\cite{spacetimegaussian, xu20244k4d}, most existing methods focus on short sequences of around 300 frames, whereas scenarios involving much longer sequences, often \textit{exceeding \textbf{1,000} frames}, remain largely underexplored. Compared with conventional short-horizon dynamic reconstruction, long-horizon streamable free-viewpoint video reconstruction poses additional challenges, since temporal errors can accumulate over time~\cite{sullivan1998rate, kamaci2005frame, wang2024v3, furecon, zheng20254dgcpro}, progressively degrading reconstruction fidelity and temporal stability under strict streaming constraints. In this setting, compact storage alone is insufficient. A practical framework must also suppress error propagation while preserving real-time efficiency and favorable RD performance~\cite{attal2023hyperreel,wu2024tetrirf}.

Existing efforts on long-horizon volumetric videos mainly follow two directions. The first focuses on offline representations for long volumetric videos. For instance, prior work~\cite{xu2024representing} models long volumetric videos using temporal Gaussian hierarchies to reduce the high memory cost of extended sequences. Although effective for offline long-sequence representation, such methods require access to the entire video during optimization and are therefore unsuitable for streamable reconstruction. 
The second direction mitigates long-term drift by partitioning the full video into multiple GOP and restarting temporal modeling within each segment~\cite{wang2024v3,chen2025motion,wang2025airgs,li2025gifstream,wang2024videorf}. While this strategy can partially alleviate error accumulation by periodically resetting temporal dependencies, it introduces substantial time and storage overhead. As a result, GOP-based approaches degrade real-time performance and overall RD efficiency, making them less suitable for low-latency immersive streaming systems. In contrast, our work aims to mitigate error propagation and maintain stable reconstruction quality for long-horizon volumetric videos under streaming constraints. Rather than partitioning the full sequence into multiple GOP, we instead pursue stable streamable reconstruction on LFVV in a fully GOP-free manner. To the best of our knowledge, this work presents the first error-resilient streamable FVV framework that maintains stable reconstruction performance for LFVV without relying on GOP partitioning.
\begin{figure*}[t]
  \centering
  \includegraphics[width=\textwidth]{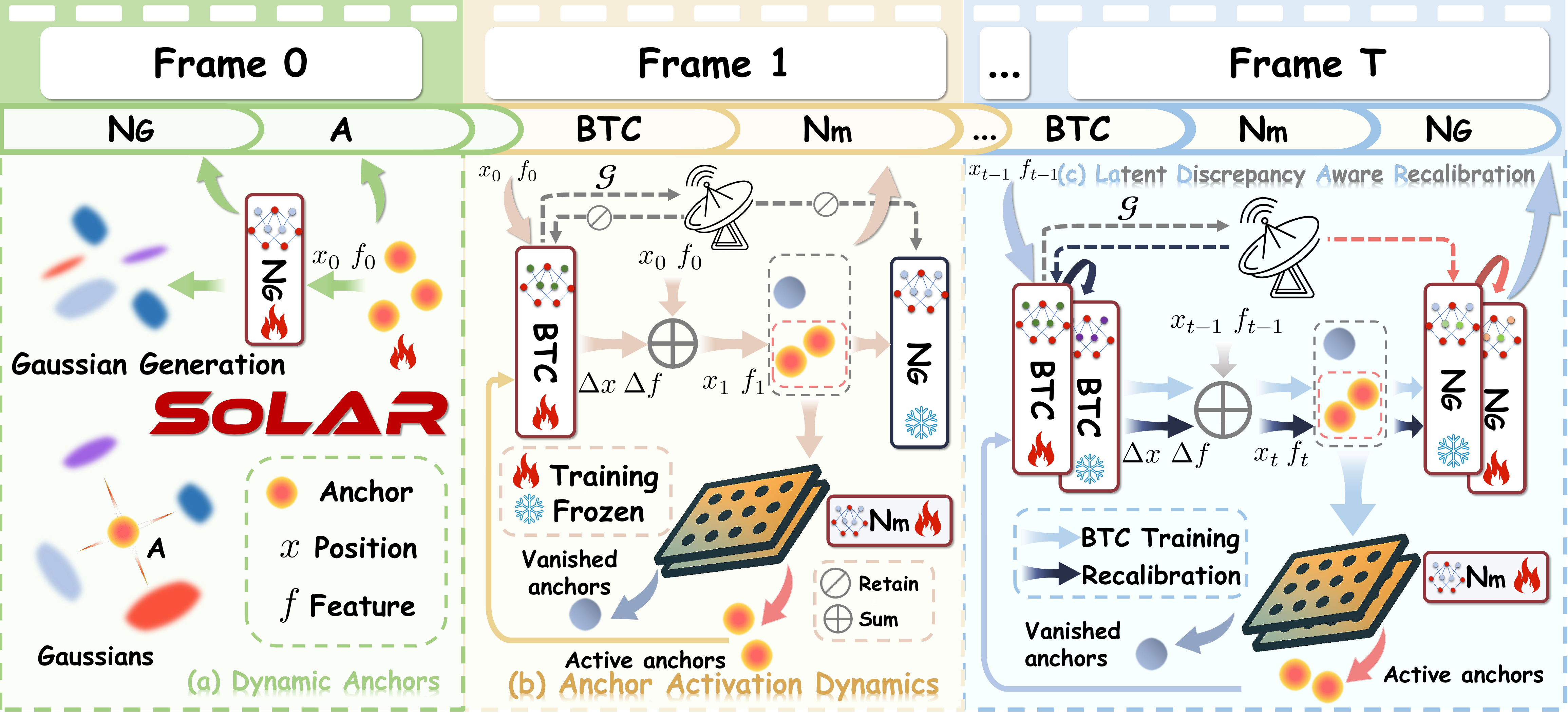}
  \caption{
    \textbf{Overview of the SoLAR framework.}
(a) \textbf{Dynamic Anchors.} Free-Viewpoint Video is represented by dynamic anchors $A$ across timesteps. $\mathbf{N}_G$ decodes Gaussians from anchor position $x$ and feature $f$. At timestep $t$, BTC predicts $\Delta x$ and $\Delta f$ from $x_{t-1}$, with its lightweight parameters encoded for transmission.
(b) \textbf{Anchor Activation Dynamics.} $\mathbf{N}_m$ generates a probability mask to split anchors into active $\mathbb{A}$ and vanished $\mathbb{V}$ (vanished anchors are masked out), with optimized $\mathbf{N}_m$ as the transmission payload.
(c) \textbf{Latent Discrepancy Aware Recalibration.} After BTC optimization, recalibration is triggered if accumulated gradient $\mathcal{G}$ exceeds the threshold: $\mathbf{N}_G$ is fine-tuned (anchor features fixed), and the updated $\mathbf{N}_G$ is transmitted. } 
  \label{fig:framework}
\end{figure*}
\section{Preliminaries}
\label{sec:Preliminaries}
\Paragraph{3D Gaussian Splatting.}
3D Gaussian Splatting (3DGS)~\cite{3DGS,ren2025octree} enables real-time free-viewpoint rendering by representing a scene as a set of explicit 3D Gaussian primitives and rendering them through differentiable splatting with tile-based rasterization. Compared with implicit neural representations, 3DGS provides an efficient and editable scene representation, where geometry and appearance are directly encoded by a collection of Gaussian primitives. Let $\mathbf{G}=\{G_i\}_{i=1}^{N}$ denote the Gaussian set, where each Gaussian is parameterized as:
\begin{equation}
G_i = (\boldsymbol{\mu}_i, \boldsymbol{s}_i, \boldsymbol{r}_i, \boldsymbol{c}_i, \alpha_i),
\end{equation}
with $\boldsymbol{\mu}_i \in \mathbb{R}^{3}$, $\boldsymbol{s}_i \in \mathbb{R}^{3}$, $\boldsymbol{r}_i \in \mathbb{R}^{4}$, $\boldsymbol{c}_i$, and $\alpha_i \in \mathbb{R}$ denoting the position, scale, rotation, appearance parameters, and opacity of the $i$-th Gaussian, respectively~\cite{3DGS,lu2024scaffold}.

\Paragraph{Rendering.}
Given a camera pose, each Gaussian is projected onto the image plane as a 2D ellipse and rasterized in a tile-based manner~\cite{3DGS,ren2025octree}. The rendered color at pixel $\boldsymbol{p}$ is computed via depth-aware alpha compositing:
\begin{equation}
C(\boldsymbol{p}) =
\sum_{i \in \mathcal{N}}
\boldsymbol{c}_i \, w_i(\boldsymbol{p})
\prod_{j=1}^{i-1}
\big(1-w_j(\boldsymbol{p})\big),
\end{equation}
where $\mathcal{N}$ denotes the set of depth-sorted Gaussians contributing to pixel $\boldsymbol{p}$, and $w_i(\boldsymbol{p})=\alpha_i \, g_i(\boldsymbol{p})$ denotes the contribution weight of the $i$-th Gaussian. The term $g_i(\boldsymbol{p})$ denotes the value of the projected 2D Gaussian at pixel $\boldsymbol{p}$ in screen space:
\begin{equation}
g_i(\boldsymbol{p}) =
\exp\!\left(
-\frac{1}{2}
(\boldsymbol{p} - \boldsymbol{\mu}'_i)^\top
(\boldsymbol{\Sigma}'_i)^{-1}
(\boldsymbol{p} - \boldsymbol{\mu}'_i)
\right).
\end{equation}
Here, $\boldsymbol{\mu}'_i$ represents the projected 2D center of the $i$-th Gaussian, and $\boldsymbol{\Sigma}'_i$ denotes the corresponding 2D covariance matrix obtained from the 3D covariance $\boldsymbol{\Sigma}_i$ through the projection Jacobian~\cite{3DGS,ren2025octree}.
Because the rasterization process is differentiable~\cite{kheradmand2025stochasticsplats}, all Gaussian parameters can be optimized end-to-end under the rendering objective~\cite{3DGS}:
\begin{equation}
\mathcal{L}_{r}
=
(1-\lambda_{\mathrm{SSIM}})\mathcal{L}_{1}
+
\lambda_{\mathrm{SSIM}}\mathcal{L}_{\mathrm{SSIM}}.
\end{equation}

\section{Methodology}
\label{sec:Methodology}
\subsection{Overview}

An overview of the \textbf{SoLAR} framework is presented in Fig.~\ref{fig:framework}, and the inter-frame transformation procedure is summarized in Algorithm~\ref{alg:inter_frame_lite}. Given a multi-view video of length $N$, \textbf{SoLAR} performs on-the-fly FVV construction and compression in a frame-wise manner. Within this framework, each frame $F_t$ is represented using Dynamic Anchors (Sec.~\ref{sec:dynamic_anchors}).

The complete scene representation is first optimized for the intra frame (I-frame), denoted by $F_0$. For each subsequent predictive frame (P-frame) $F_t$, an efficient inter-frame transformation is applied to evolve the representation of the previous frame $F_{t-1}$ into $F_t$. Specifically, a compact BTC is introduced to model anchor transformations between adjacent frames, thereby reducing temporal redundancy. During the optimization of $F_t$, Anchor Activation Dynamics is employed  to dynamically activate informative anchors while suppressing redundant ones (Sec.~\ref{sec:Anchor_Activation_Dynamics}). Furthermore, when pronounced discrepancies emerge in the latent representation, Latent Discrepancy Aware Recalibration is applied to recalibrate $F_t$ and mitigate error propagation (Sec.~\ref{sec:Latent_Discrepancy_Aware_Recalibration}). Finally, to achieve favorable rate-distortion performance, the entire training procedure is formulated as a joint optimization problem over model size (rate) and reconstruction quality (distortion) (Sec.~\ref{Sec:Sparse-Aware RD Optimization}).
\subsection{Dynamic Anchors}
\label{sec:dynamic_anchors}

In the proposed dynamic-anchor-based volumetric video representation framework, FVV is represented by a set of dynamic anchors $A$ that evolve over time. To model the temporal evolution of anchors in a compact yet expressive manner, we employ BTC to capture frame-to-frame anchor variation, and further apply the pruning strategy proposed in~\cite{ma2025hashgridfeaturepruning} to BTC to maintain model compactness. Each anchor is parameterized by a spatial location $x \in \mathbb{R}^{3}$, a latent feature vector $f \in \mathbb{R}^{D}$, and an anchor scaling factor $l \in \mathbb{R}^{3}$. To generate $k$ Gaussians from each anchor, the Gaussian attribute network $\mathbf{N}_G$ learns a mapping from $f$ to the Gaussian attributes $\mathcal{A}$~\cite{lu2024scaffold,tang2025compressing}:
\begin{align}
\{o_i, c_i, r_i, s_i, \alpha_i\}_{i=0}^{k-1}
&= \mathbf{N}_G(f, \overrightarrow{d}_c), \\
\{\mu_i\}_{i=0}^{k-1} &= x + \{o_i\}_{i=0}^{k-1} \cdot l,
\label{eq:mlp_prediction}
\end{align}
where $\overrightarrow{d}_c$ denotes the viewing direction, and $o_i \in \mathbb{R}^{3}$, $c_i \in \mathbb{R}^{3}$, $r_i \in \mathbb{R}^{4}$, $s_i \in \mathbb{R}^{3}$, $\alpha_i \in \mathbb{R}$, and $\mu_i \in \mathbb{R}^{3}$ denote the offset, color, rotation quaternion, scale, opacity, and location of the $i$-th Gaussian, respectively. The attributes of the $i$-th Gaussian are defined as $\mathcal{A}_i = \{\mu_i, c_i, r_i, s_i, \alpha_i\}$~\cite{3DGS,tang2025compressing,lu2024scaffold}.

At timestep $t$, BTC propagates the anchor state from the previous timestep by jointly modeling the spatial evolution and latent feature variation of each anchor~\cite{tang2025compressing}. Instead of storing dense frame-specific parameters for all Gaussians, BTC maintains a compact transformation representation that operates directly in the anchor domain. This design preserves the compactness of the anchor-based representation while enabling temporally evolving geometry and appearance to be captured through lightweight anchor-level updates.

Specifically, BTC first extracts a compact transformation representation from the predecessor anchor state. Based on this shared transformation context, anchor-BTC $\mathbf{BTC}_x$ jointly predicts a spatial update together with two scaling coefficients, while feature-BTC $\mathbf{BTC}_f$ predicts the feature residual. In addition to the update itself, we explicitly introduce scaling prediction to account for the magnitude variation of neural network outputs. This allows the predicted updates to be adaptively rescaled before being applied, thereby improving the stability and flexibility of the transformation modeling:
\begin{align}
[\Delta \tilde{x}_t,\gamma_t^{x},\gamma_t^{f}] &= \mathbf{BTC}_x(x_{t-1}), \\
\Delta \tilde{f}_t &= \mathbf{BTC}_f(x_{t-1}),
\end{align}
where $\Delta \tilde{x}_t \in \mathbb{R}^{3}$ and $\Delta \tilde{f}_t \in \mathbb{R}^{D}$ denote the candidate motion update and feature residual, respectively, and $\gamma_t^{x}, \gamma_t^{f} \in \mathbb{R}$ denote the corresponding learned scaling factors. The final anchor updates are then obtained via scale modulation:
\begin{align}
\Delta x_t &= \gamma_t^{x} \cdot \Delta \tilde{x}_t,
\qquad
x_t = x_{t-1} + \Delta x_t, \\
\Delta f_t &= \gamma_t^{f} \cdot \Delta \tilde{f}_t,
\qquad
f_t = f_{t-1} + \Delta f_t.
\label{eq:btc_update}
\end{align}

This formulation offers two key benefits. First, anchor-BTC predicts the spatial update and both scaling factors jointly, allowing a shared transformation context to coordinate geometric evolution and feature adaptation in a more unified manner. Second, by decomposing anchor evolution into a residual component and a learned scale term, BTC can adaptively modulate the magnitude of both motion and feature updates according to local dynamic complexity. As a result, the proposed dynamic anchor modeling mechanism enhances temporal adaptability under heterogeneous non-rigid dynamics while preserving the compactness and scalability required for long-horizon streamable free-viewpoint video reconstruction.

\subsection{Anchor Activation Dynamics}
\label{sec:Anchor_Activation_Dynamics}

Although BTC effectively models coordinate shifts, it assumes an anchor set with fixed cardinality, which limits its adaptability under \textit{topological changes}, such as the occlusion of previously visible objects or the emergence of new content~\cite{sun20243dgstream,jiang2025topology,zhang2025evolvinggs}. Under such dynamics, retaining all anchors is suboptimal: some anchors may cease to correspond to informative scene regions, while newly important regions may require additional representational support~\cite{lee2024compact,lightgaussian,liu2025maskgaussian}. To overcome this limitation, we introduce \emph{Anchor Activation Dynamics}, a mechanism that endows the anchor set with frame-adaptive selectivity by preserving informative anchors and suppressing redundant ones.
Specifically, for each anchor with latent feature $f$ and spatial location $x$, the anchor mask network $\mathbf{N}_m$ predicts a scalar significance score $m_x \in \mathbb{R}$, which is converted into a gating variable that modulates the scale and opacity attributes of the associated Gaussians:
\begin{align}
&m_x = \mathbf{N}_m(f, \,x), \\
&M_x = \operatorname*{SG}\!\left(\mathbb{I}[\sigma(m_x) > \epsilon_m] - \sigma(m_x)\right) + \sigma(m_x), \\
&\{\hat{s}_i, \hat{\alpha}_i\}_{i=0}^{k-1} = M_x \cdot \{s_i, \alpha_i\}_{i=0}^{k-1},
\end{align}
where $\epsilon_m$ denotes the mask threshold, $\operatorname*{SG}(\cdot)$ denotes the stop-gradient operator~\cite{wang2024end}, $\mathbb{I}[\cdot]$ denotes the indicator function, and $\sigma(\cdot)$ denotes the sigmoid function. Through this gating operation, low-significance anchors are attenuated toward zero contribution, whereas high-significance anchors preserve their geometric attributes. The formulation employs the Straight-Through Estimator, so that anchor activation follows a thresholded binary decision in the forward pass while remaining differentiable through the sigmoid branch in the backward pass, thereby enabling end-to-end learning of the mask network.

\Paragraph{Activation Dynamics.}
A key aspect of this design is that anchor activation is updated at every timestep, allowing anchor states to evolve with scene topology instead of being fixed once initialized. During optimization, all anchors are refined jointly, enabling the learned mask scores to separate anchors that remain structurally relevant from those that have become uninformative. According to these scores, anchors are partitioned into active anchors $\mathbb{A}$ and vanished anchors $\mathbb{V}$:
\begin{align}
\mathbb{A} &= \{A_i \mid \sigma(m_x) \in (\epsilon_{m}, 1] \land i \in [1, |A|]\}, \\
\mathbb{V} &= \{A_i \mid \sigma(m_x) \in [0, \epsilon_{m}] \land i \in [1, |A|]\}.
\end{align}

During rendering, anchors in $\mathbb{V}$ are frozen and masked out, while anchors in $\mathbb{A}$ remain active for final image formation. This selective update strategy concentrates representational capacity on anchors that remain topologically relevant to the current frame, while avoiding unnecessary updates to anchors that no longer correspond to meaningful scene content. After optimization, the weights of $\mathbf{N}_m$ are encoded and transmitted as a warm-start initialization for the next frame. Since adjacent frames typically share substantial scene structure, this initialization naturally encourages temporally coherent anchor activation across frames and provides a more stable starting point for subsequent optimization.
\subsection{Latent Discrepancy Aware Recalibration}
\label{sec:Latent_Discrepancy_Aware_Recalibration}

iFVC~\cite{tang2025compressing} employs a shared Gaussian attribute network $\mathbf{N}_G$ for all P-frames $F_t$. This network is trained at timestep $0$ and remains fixed throughout the sequence. 
Conceptually, training $\mathbf{N}_G$ induces a mapping $\mathcal{K}: f \mapsto \mathcal{A}$ from the anchor latent feature space to the Gaussian attribute space, such that each frame relies on its own latent feature $f$ to decode $\mathcal{A}$. 
While this static design is highly storage-efficient, it implicitly assumes that the latent features of later frames remain compatible with the feature-to-attribute correspondence learned at the initial timestep. In practice, this assumption becomes increasingly fragile, as anchor latent features evolve with changes in scene content, viewpoint, and motion over time. Consequently, the fixed mapping $\mathcal{K}$ encoded in $\mathbf{N}_G$ can become progressively mismatched to the time-varying query features, leading to increasingly unreliable Gaussian attribute prediction.
This mismatch is particularly detrimental in predictive streaming, where errors at the current frame directly affect the representation quality of subsequent frames. Once the decoded Gaussian attributes deviate from the latent representation at timestep $t$, the resulting prediction bias can propagate forward and progressively degrade later-frame reconstruction. Consequently, even modest discrepancies introduced at early stages may accumulate over time and ultimately lead to severe degradation in long-horizon video scenarios.

A straightforward way to alleviate this mismatch is to update the mapping $\mathcal{K}$ for each frame, allowing the Gaussian attribute network to adapt to the latent feature distribution at the current timestep. However, directly reconstructing or updating $\mathbf{N}_G$ in a frame-wise manner would introduce substantial storage overhead, since the frame-specific network updates must also be stored and transmitted to the decoder. Therefore, although such updates may reduce the discrepancy between latent features and Gaussian attributes, the additional parameter cost would compromise the overall rate-distortion efficiency.
To address this trade-off, we propose \textit{Latent Discrepancy Aware Recalibration} (LaDAR), which aims to adapt the learned feature-to-attribute correspondence without relying on explicit frame-wise updates of $\mathbf{N}_G$. Specifically, LaDAR identifies the discrepancy between the evolving latent feature $f$ and the correspondence $\mathcal{K}$ encoded in the fixed Gaussian attribute network $\mathbf{N}_G$. In this way, the decoded Gaussian attributes can remain consistently aligned with the current latent representation, thereby mitigating temporal mismatch and suppressing error accumulation in predictive streaming.
Motivated by the bit allocation theory in traditional video coding~\cite{li2016lambda}, which achieves optimal RD performance by allocating more bitrate to critical frames, we first establish a theoretical foundation for RD optimization under rate-control constraints, and then apply this to dynamic-anchor-based volumetric video representations. For conciseness, the main results are presented below, with the full derivation deferred to the supplementary material.
\begin{blackclaim}
\begin{proposition}
\label{main_paper_pro:important frames should be encoded with smaller distortion}
To achieve global RD optimality, more critical frames should be allocated lower distortion levels.
\end{proposition}
\end{blackclaim}

A frame is considered more critical if its reconstruction quality exerts a stronger influence on subsequent frames. We refer to this downstream influence as the \textit{temporal importance} of the frame. In our dynamic-anchor-based volumetric video representation, we quantify temporal importance through discrepancy propagation. Specifically, the discrepancy-based temporal importance of each frame is defined as follows:

\begin{definition}[Discrepancy-based Importance]
\begin{equation}
\theta_t = \frac{\partial \left( \sum_{i=t+1}^{N} \mathcal{D}_i \right)}{\partial \mathcal{D}_t},
\label{eq:influence_discrepancy}
\end{equation}
where $\mathcal{D}_t$ denotes the discrepancy between $f$ and $\mathbf{N}_G$ at timestep $t$, and $\theta_t$ quantifies the importance of $F_t$. Specifically, $\theta_t$ measures how a perturbation in the discrepancy at frame $t$ affects the cumulative discrepancy of future frames. A larger $\theta_t$ thus indicates that the current frame exerts a stronger temporal influence, meaning that reducing its discrepancy leads to a greater downstream reduction in sequence-level distortion.
\end{definition}
Although this definition provides a principled characterization of frame importance, directly estimating the influence of the current frame on \emph{all} subsequent frames is computationally intractable in a streaming setting. Exact evaluation would require tracking how the discrepancy at the current frame propagates through all future frames, while such future information is unavailable at the current timestep under the on-the-fly streaming paradigm~\cite{sun20243dgstream,tang2025compressing}. To address this limitation, we introduce a tractable online surrogate based on the gradient information generated during the BTC-based transformation of the latent feature $f$. The effectiveness of this gradient-based surrogate is empirically validated in Sec.~\ref{sec:experiment:correlation}.
The key intuition is that, when the transformed latent feature becomes increasingly mismatched with the fixed correspondence, optimizing $\mathbf{BTC}_f$ requires stronger corrective updates, which are reflected by larger gradients during BTC-based feature transformation. Therefore, the gradient signal serves as an efficient discrepancy indicator: larger gradients suggest a more pronounced mismatch between $f$ and $\mathbf{N}_G$, which may degrade Gaussian prediction and exacerbate error accumulation.

To improve the robustness of the gradient-based discrepancy indicator, we maintain an exponential moving average (EMA) of the gradients associated with $\mathbf{BTC}_f$. This design suppresses short-term fluctuations caused by stochastic viewpoint selection while retaining the underlying optimization tendency. As a result, it provides a reliable estimate of the discrepancy state near convergence. Specifically, the EMA statistic $\mathcal{G}$ is updated with a smoothing coefficient $\alpha_{\mathcal{D}}$ set to $0.3$ by default:
\begin{equation}
\mathcal{G}_i = \alpha_\mathcal{D} \cdot \mathcal{G}_{i-1} + (1 - \alpha_\mathcal{D}) \cdot g_i, \quad i = 1,\dots,T_{\text{BTC}},
\label{gradient static}
\end{equation}
where $g_i$ and $\mathcal{G}_i$ denote the gradient of $\mathbf{BTC}_f$ and its exponential moving average at step $i$, respectively, and $T_{\text{BTC}}$ denotes the total number of training steps for BTC. The resulting statistic $\mathcal{G}_{T_{\text{BTC}}}$ therefore serves as a compact indicator of the underlying latent discrepancy.

\Paragraph{Recalibration.}
After the BTC training phase, recalibration of $\mathbf{N}_G$ is activated only when the final gradient statistic $\mathcal{G}_{T_{\text{BTC}}}$ exceeds a predefined discrepancy threshold $\epsilon_\mathcal{D}$. This event-driven criterion prevents unnecessary network updates and ensures that recalibration is performed only when the observed discrepancy is sufficiently large to suggest that the fixed correspondence $\mathcal{K}$ is no longer well aligned with the current latent distribution. During recalibration, $\mathbf{N}_G$ is fine-tuned while the anchor features are kept frozen. This design explicitly updates the correspondence $\mathcal{K}$ encoded in $\mathbf{N}_G$, rather than re-optimizing the latent representation already learned for the current frame. The recalibrated $\mathbf{N}_G$ is then stored and transmitted to the decoder, enabling subsequent frames to decode Gaussian attributes with an updated network that better matches the evolving feature distribution. Since recalibration is selectively invoked and $\mathbf{N}_G$ is compact, this procedure introduces only limited computational and transmission overhead, thereby preserving the storage efficiency of the framework.
\subsection{Sparse-Aware RD Optimization}
\label{Sec:Sparse-Aware RD Optimization}

To jointly balance reconstruction quality, coding efficiency, and anchor compactness at time $t$, we formulate the training objective $\mathcal{L}^{(t)}$ as a weighted combination of three complementary terms: the rendering loss $\mathcal{L}_r^{(t)}$, the entropy loss $\mathcal{L}_e^{(t)}$, and the sparsity loss $\mathcal{L}_s^{(t)}$:
\begin{equation}
\mathcal{L}^{(t)} = \mathcal{L}_r^{(t)} + \lambda_e \mathcal{L}_e^{(t)} + \lambda_s \mathcal{L}_s^{(t)},
\end{equation}
where $\lambda_e$ and $\lambda_s$ are trade-off coefficients. Specifically, $\mathcal{L}_r^{(t)}$ enforces rendering accuracy, $\mathcal{L}_e^{(t)}$ constrains the number of transmitted bits, and $\mathcal{L}_s^{(t)}$ encourages a more compact anchor configuration. By optimizing these terms jointly, sparse anchor selection is incorporated directly into the RD objective itself, rather than imposed afterward as a separate post hoc constraint.

The rendering loss~\cite{tang2025compressing} is defined as:
\begin{equation}
\mathcal{L}_r^{(t)} =(1-\lambda_{\text{SSIM}}) \mathcal{L}_1 + \lambda_{\text{SSIM}} \mathcal{L}_{\text{SSIM}},
\end{equation}
where $\mathcal{L}_1$ denotes the pixel-wise $\ell_1$ loss, $\mathcal{L}_{\text{SSIM}}$ represents the structural similarity loss, and $\lambda_{\text{SSIM}}$ balances the contribution of the structural similarity term, which is set to $0.2$ by default.

For the entropy term, we follow prior work~\cite{tang2025compressing} and estimate the bit consumption of BTC from the occurrence probability $p_B$ of the binary symbol $+1$. Specifically, the generated BTC symbols are modeled with a Bernoulli distribution, under which an observed $+1$ incurs a code length of $-\log_2(p_B)$ and an observed $-1$ incurs $-\log_2(1-p_B)$. The entropy loss at time $t$ is therefore defined as:
\begin{equation}
\mathcal{L}_e^{(t)} = C_+ \times (-\log_2(p_B)) + C_- \times (-\log_2(1 - p_B)),
\end{equation}
where $C_+$ and $C_-$ denote the total counts of symbols $+1$ and $-1$, respectively. 

To further encourage compactness, we introduce a sparsity-aware regularization term:
\begin{equation}
\mathcal{L}_s^{(t)} = \frac{1}{N_a} \sum_{n=1}^{N_a} \sigma(m_n),\quad N_a = |A|,
\end{equation}
where $N_a$ is the total number of anchors. Instead of penalizing hard binary decisions, this term operates on the mean soft activation of anchors, which preserves compatibility with gradient-based optimization. By suppressing unnecessarily high activation scores, it steers the model toward using fewer active anchors, thereby promoting a compact representation while retaining anchors that remain beneficial for rendering.
\renewcommand{\algorithmicrequire}{\textbf{Input:}}
\renewcommand{\algorithmicensure}{\textbf{Output:}}
\newcommand{\algorithmichyperparameter}{\textbf{Hyperparameters:}}
\newcommand{\HYPERPARAMETER}{\item[\algorithmichyperparameter]}
\begin{algorithm}[t]
\caption{Inter-frame Transformation Procedure of SoLAR}
\label{alg:inter_frame_lite}
\begin{algorithmic}[1]
\REQUIRE $\mathbf{I}_t$, $A_{t-1}$, $\mathbf{N}_m^{t-1}$, $\mathbf{N}_G$, random initialization $\Delta$,\\ rasterizer $\mathbf{R}$
\ENSURE $A_t$, $\mathbf{N}_m^t$, $\mathbf{BTC}_x$, $\mathbf{BTC}_f$, $\mathbf{N}_G$
\HYPERPARAMETER $T_{\text{BTC}}$, $T_{\text{R}}$,$\alpha_{\mathcal D}$,  $\epsilon_{\mathcal D}$,$\epsilon_m$,$\lambda_e$,$\lambda_s$

\STATE Initialize $A_t \leftarrow A_{t-1}$, $\mathbf{BTC}_x,\mathbf{BTC}_f \leftarrow \Delta$, $\mathbf{N}_m^t \leftarrow \mathbf{N}_m^{t-1}$ or $\Delta$, and $\mathcal{G}_0 \leftarrow 0$
\FOR{$i=1$ to $T_{\text{BTC}}$}
    \STATE $[\Delta \tilde{x}_t,\gamma_t^{x},\gamma_t^{f}] = \mathbf{BTC}_x(x_{t-1}),
\Delta \tilde{f}_t = \mathbf{BTC}_f(x_{t-1})$
    \STATE $\Delta x_t = \gamma_t^{x} \cdot \Delta \tilde{x}_t,
\quad x_t = x_{t-1} + \Delta x_t, $
    \STATE  $\Delta f_t = \gamma_t^{f} \cdot \Delta \tilde{f}_t,
\quad f_t = f_{t-1} + \Delta f_t$
    \STATE $\mathcal{A} \leftarrow \mathbf{N}_G(f_t, \overrightarrow{d}_c)$, \quad $m \leftarrow \mathbf{N}_m^t(x_t,f_t)$
    \STATE $M \leftarrow \operatorname*{SG}\!\left(\mathbb{I}[\sigma(m)>\epsilon_m]-\sigma(m)\right)+\sigma(m)$
    \STATE $\mathbf{I}_t' \leftarrow \mathbf{R}(M\cdot\mathcal{A})$
    \STATE $\mathcal{L}^{(t)} \leftarrow \mathcal{L}_r^{(t)}+\lambda_e\mathcal{L}_e^{(t)}+\lambda_s\mathcal{L}_s^{(t)}$
    \STATE $g_i=\left\|\nabla_{\mathbf{BTC}_f}\mathcal{L}^{(t)}\right\|_2$ 
    \STATE
    $\mathcal{G}_i \leftarrow \alpha_\mathcal{D}\mathcal{G}_{i-1}+(1-\alpha_\mathcal{D})g_i$
    \STATE Update $\mathbf{BTC}_x,\mathbf{BTC}_f,\mathbf{N}_m^t$
\ENDFOR
\IF{$\mathcal{G}_{T_{\text{BTC}}}>\epsilon_\mathcal{D}$}
    \FOR{$i=1$ to $T_{\text{R}}$}
        \STATE Render with $(x_t,f_t,\mathbf{N}_m^t,\mathbf{N}_G)$, compute $\mathcal{L}_r^{(t)}$
        \STATE Update $\mathbf{N}_G$
    \ENDFOR
\ENDIF
\RETURN $A_t$, $\mathbf{N}_m^t$, $\mathbf{BTC}_x$, $\mathbf{BTC}_f$, $\mathbf{N}_G$
\end{algorithmic}
\end{algorithm}

\section{Experiments}
\label{section:Experiments}
\begin{table*}[!t]
\centering
\caption{
\textbf{Quantitative comparison on Meeting Room and N3DV datasets.}
A dash (--) indicates that the corresponding metric is not reported in the original work.
\textbf{SoLAR-lite} denotes the lightweight version of \textbf{SoLAR}.
}
\label{tab:meetroom_n3dv}

\scriptsize
\setlength{\tabcolsep}{3.8pt}
\renewcommand{\arraystretch}{0.96}

\resizebox{\textwidth}{!}{
\begin{tabular}{c l cccc cccc}
\toprule
\multirow{2}{*}{Category} &\multirow{2}{*}{\centering Method}
& \multicolumn{4}{c}{Meeting Room}
& \multicolumn{4}{c}{N3DV} \\
\cmidrule(lr){3-6} \cmidrule(lr){7-10}
& 
& PSNR(dB)$\uparrow$ & SSIM$\uparrow$ & Storage(MB)$\downarrow$ & Render(FPS)$\uparrow$
& PSNR(dB)$\uparrow$ & SSIM$\uparrow$ & Storage(MB)$\downarrow$ & Render(FPS)$\uparrow$ \\
\midrule

\multirow{2}{*}{Offline}
& Swift4D (ICLR 2025)
& 32.05 & -- & 0.13 & --
& 32.23 & -- & 0.40 & 125 \\
& SplineGS (ICLR 2025)
& -- & -- & -- & --
& 32.60 & 0.950 & -- & 76 \\
\midrule

\multirow{10}{*}{Online}
& 3DGStream (CVPR 2024)
& 29.30 & -- & 4.10 & 260
& 31.35 & 0.948 & 7.80 & 245 \\
& 4DGC (CVPR 2025)
& 28.08 & 0.922 & 0.42 & 213
& 31.58 & 0.943 & 0.50 & 168 \\
& 4DGCPro (NeurIPS 2025)
& -- & -- & -- & --
& 31.64 & 0.944 & 0.64 & -- \\
& ComGS-l (NeurIPS 2025)
& -- & -- & -- & --
& 32.12 & -- & 0.11 & 147 \\
& iFVC (AAAI 2025)
& 32.38 & -- & 0.09 & 157
& 32.32 & -- & 0.09 & 126 \\
& QUEEN-l (NeurIPS 2024)
& -- & -- & -- & --
& 32.19 & 0.946 & 0.75 & 248 \\
& D-FCGS (AAAI 2026)
& 30.97 & 0.950 & 0.09 & \cellbest{288}
& 32.02 & 0.951 & 0.18 & 215 \\
& HiCoM (NeurIPS 2024)
& 26.73 & -- & 0.60 & \cellsecond{284}
& 32.08 & 0.953 & 0.69 & \cellbest{255} \\
& Recon-GS (NeurIPS 2025)
& 30.84 & -- & 0.30 & 256
& 32.66 & \cellbest{0.957} & 0.44 & \cellsecond{250} \\
& \textbf{SoLAR (Ours)}
& \cellbest{33.62} & \cellbest{0.962} & \cellsecond{0.05} & 127
& \cellbest{33.43} & \cellsecond{0.954} & \cellsecond{0.06} & 160 \\
& \textbf{SoLAR-lite (Ours)}
& \cellsecond{33.04} & \cellsecond{0.958} & \cellbest{0.02} & 182
& \cellsecond{32.93} & 0.951 & \cellbest{0.03} & 163 \\
\bottomrule
\end{tabular}
}
\end{table*}
\begin{table}[t]
\centering
\caption{\textbf{Quantitative comparison on Bar.} $^{\dag}$ denotes GOP partitioning.}
\label{tab:bar_table}

\scriptsize
\setlength{\tabcolsep}{1.8pt}
\renewcommand{\arraystretch}{0.95}

\begin{tabularx}{\linewidth}{M *{6}{C}}
\toprule
Method 
& PSNR$\uparrow$ 
& SSIM$\uparrow$ 
& LPIPS$\downarrow$ 
& Storage$\downarrow$ 
& Train$\downarrow$ 
& Render$\uparrow$ \\
& (dB) & & & (MB) & (min) & (FPS) \\
\midrule
3DGStream & 26.78 & \cellsecond 0.853 & \cellsecond 0.162 & 3.80 & 1.127 & 114 \\
4DGC & 28.11 & 0.800 & 0.167 & 1.22 & \cellsecond 0.867 & 156 \\
HiCoM & 23.24 & 0.676 & 0.380 & 0.89 & 0.879 & 163 \\
Recon-GS & 27.43 & 0.803 & 0.250 & \cellsecond 0.32 & 0.936 & 176 \\
iFVC & 28.03 & 0.842 & 0.194 & 0.42 & 0.906 & \cellbest 213 \\
iFVC$^\dag$ & \cellsecond 28.13 & 0.852 & 0.183 & 0.42 & 1.134 & 176 \\
\textbf{SoLAR} & \cellbest{32.45} & \cellbest{0.893} & \cellbest{0.145} & \cellbest{0.05} & \cellbest 0.852 & \cellsecond 180 \\
\bottomrule
\end{tabularx}
\end{table}
\begin{table}[t]
\centering
\caption{\textbf{Quantitative comparison on AvatarRex-lbn1.}}
\label{tab:quantitative_result_lbn1}

\scriptsize
\setlength{\tabcolsep}{2pt}
\renewcommand{\arraystretch}{0.92}

\begin{tabularx}{\linewidth}{M *{5}{C}}
\toprule
Method & PSNR(dB)$\uparrow$ & SSIM$\uparrow$ & LPIPS$\downarrow$ & Storage(MB)$\downarrow$ & Render(FPS)$\uparrow$ \\
\midrule
3DGStream & \cellsecond 28.11 & 0.935 & \cellsecond 0.081 & 3.80 & 321 \\
4DGC & 24.53 & 0.885 & 0.110 & 1.41 & 166 \\
HiCoM & 25.74 & 0.864 & 0.094 & 0.10 & \cellbest 371 \\
Recon-GS & 24.48 & 0.889 & 0.102 & \cellsecond 0.08 & 184 \\
iFVC & 27.32 & \cellsecond 0.938 & 0.089 & 0.10 & 201 \\
\textbf{SoLAR} & \cellbest{32.28} & \cellbest{0.965} & \cellbest{0.067} & \cellbest{0.05} & \cellsecond 346 \\
\bottomrule
\end{tabularx}
\end{table}
\begin{table}[t]
\centering
\caption{\textbf{Quantitative comparison on AvatarRex-lbn2.}}
\label{tab:quantitative_result_lbn2}

\scriptsize
\setlength{\tabcolsep}{2pt}
\renewcommand{\arraystretch}{0.92}

\begin{tabularx}{\linewidth}{M *{5}{C}}
\toprule
Method & PSNR(dB)$\uparrow$ & SSIM$\uparrow$ & LPIPS$\downarrow$ & Storage(MB)$\downarrow$ & Render(FPS)$\uparrow$ \\
\midrule
3DGStream & 24.97 & 0.906 & 0.090 & 3.80 & 289 \\
4DGC & 21.48 & 0.858 & 0.120 & 1.41 & 196 \\
HiCoM & 23.49 & 0.853 & 0.098 & 0.12 & \cellbest 383 \\
Recon-GS & 22.51 & 0.828 & 0.135 & 0.10 & 203 \\
iFVC & \cellsecond 26.06 & \cellbest 0.930 & \cellsecond 0.080 & \cellsecond 0.09 & \cellsecond 344 \\
\textbf{SoLAR} & \cellbest{27.46} & \cellsecond 0.928 & \cellbest{0.071} & \cellbest{0.08} & 228 \\
\bottomrule
\end{tabularx}
\end{table}
\begin{table}[t]
\centering
\caption{\textbf{Quantitative comparison on AvatarRex-zxc.}}
\label{tab:quantitative_result_zxc}

\scriptsize
\setlength{\tabcolsep}{2pt}
\renewcommand{\arraystretch}{0.92}

\begin{tabularx}{\linewidth}{M *{5}{C}}
\toprule
Method & PSNR(dB)$\uparrow$ & SSIM$\uparrow$ & LPIPS$\downarrow$ & Storage(MB)$\downarrow$ & Render(FPS)$\uparrow$ \\
\midrule
3DGStream & 23.17 & 0.895 & 0.104 & 3.80 & 222 \\
4DGC & 20.66 & 0.837 & 0.133 & 1.40 & 167 \\
HiCoM & 20.67 & 0.756 & 0.132 & 0.17 & \cellsecond 276 \\
Recon-GS & 20.65 & 0.791 & 0.126 & 0.10 & 146 \\
iFVC & \cellsecond 24.06 & \cellsecond 0.928 & \cellsecond 0.091 & \cellsecond 0.07 & 101 \\
\textbf{SoLAR} & \cellbest{27.63} & \cellbest{0.947} & \cellbest{0.078} & \cellbest{0.06} & \cellbest{326} \\
\bottomrule
\end{tabularx}
\end{table}
\subsection{Experiment Setup}
\label{sec:experiment:Experiment Setup}
\Paragraph{Datasets.} Experiments are conducted on four representative and widely used real-world dynamic scene datasets:
\begin{itemize}
    \item Neural 3D Video dataset (N3DV)~\cite{n3dv} comprises 6 dynamic scenes. Each scene contains 18 to 21 multi-view videos, with each video spanning 300 frames captured at a resolution of $2704 \times 2028$. Following prior works~\cite{tang2025compressing,wu2025swift4d,sun20243dgstream}, we downsample the videos by a factor of two.
    \item Meeting Room dataset (Meeting Room)~\cite{streamrf} consists of three dynamic scenes from diverse real-world environments. Each scene is recorded by 13 synchronized multi-view videos, with each video spanning 300 frames captured at a resolution of $1280 \times 720$. 
    \item Bar dataset (Bar)~\cite{xu2024representing} consists of high-resolution video recordings captured at 60 FPS in 4K using a synchronized array of 22 cameras, containing over 3,000 frames. This dataset is particularly challenging due to its long duration and the large magnitude of scene motion. 
    
    \item AvatarRex dataset (AvatarRex)~\cite{zheng2023avatarrex} comprises three multi-view video sequences captured with a synchronized array of 16 well-calibrated RGB cameras at 30 FPS. Each sequence is recorded at a resolution of $1500 \times 2048$, with lengths ranging from 1{,}800 to 2{,}000 frames. 
    
\end{itemize}
To ensure a fair comparison, we follow prior work~\cite{tang2025compressing,sun20243dgstream,wu20244d,furecon,gao2024hicom} by holding out one viewpoint for testing and using the remaining viewpoints for training.

\Paragraph{Baselines.} We compare \textbf{SoLAR} against a set of recent and representative state-of-the-art methods, covering both offline and online approaches. The offline baselines include Swift4D~\cite{wu2025swift4d} and SplineGS~\cite{yoonsplinegs}, while the online baselines include 3DGStream~\cite{sun20243dgstream}, iFVC~\cite{tang2025compressing}, Recon-GS~\cite{furecon}, QUEEN~\cite{queen}, D-FCGS~\cite{zhang2025d}, 4DGCPro~\cite{zheng20254dgcpro}, ComGS~\cite{chen2025motion}, 4DGC~\cite{hu20254dgc}, and HiCoM~\cite{gao2024hicom}. For each dataset, we follow a consistent comparison protocol. Specifically, we use the results reported in the original papers  when they are available for the same dataset. When such results are unavailable, we reproduce the corresponding methods using their publicly released code under the same experimental settings. If a method provides neither reported results nor a public implementation for a given dataset, we leave the corresponding entries blank.

\Paragraph{Metrics.} To evaluate different methods, we report commonly used metrics, including PSNR (dB), SSIM~\cite{wang2004image}, LPIPS~\cite{zhang2018unreasonable}, storage (MB), training time (min), and rendering speed (FPS). All metrics are averaged over the entire sequence. Experiments are repeated over multiple runs to ensure statistical reliability.

\Paragraph{Implementation Details.} The default settings are described as follows. The framework begins with sparse points reconstructed by Structure-from-Motion (SfM) at timestep 0. We train frame $F_0$ for 15K steps using a context model~\cite{chen2024hac} to compress anchor attributes. For each subsequent frame $F_t$, we train BTC for 500 steps. The thresholds $\epsilon_m$ and $\epsilon_\mathcal{D}$ are set to 0.01 and 0.002, respectively. The weight parameters $\lambda_e$ and $\lambda_s$ are set to 0.004 and 0.01, respectively. The implementation of BTC and $\mathbf{N}_G$ follows prior work~\cite{tang2025compressing}. After BTC training, if recalibration is required, an additional 200 fine-tuning steps are applied to $\mathbf{N}_G$. We use the Adam optimizer with a learning rate of $5 \times 10^{-3}$. Experiments are conducted on an NVIDIA RTX L20 GPU. We also implement a lightweight variant, denoted as \textit{SoLAR-lite}, by reducing the size of BTC and $\mathbf{N}_G$.
\begin{figure*}[t]
\centering
\setlength{\tabcolsep}{0pt}

\begin{minipage}[t]{0.168\textwidth}
    \centering
    \parbox[c][0.42cm][c]{\linewidth}{\centering \small Ground truth}
    \includegraphics[width=\linewidth,keepaspectratio]{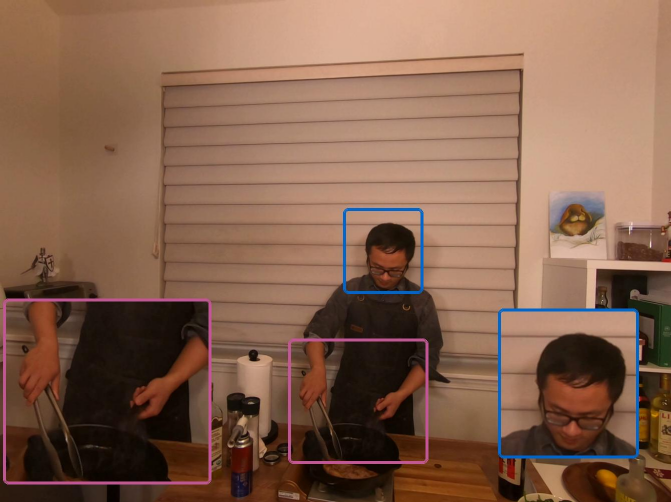}
\end{minipage}%
\begin{minipage}[t]{0.168\textwidth}
    \centering
    \parbox[c][0.42cm][c]{\linewidth}{\centering \small 3DGStream}
    \includegraphics[width=\linewidth,keepaspectratio]{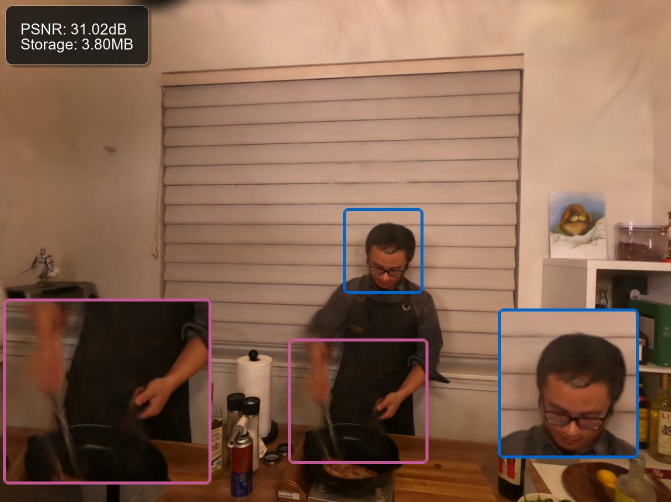}
\end{minipage}%
\begin{minipage}[t]{0.168\textwidth}
    \centering
    \parbox[c][0.42cm][c]{\linewidth}{\centering \small 4DGC}
    \includegraphics[width=\linewidth,keepaspectratio]{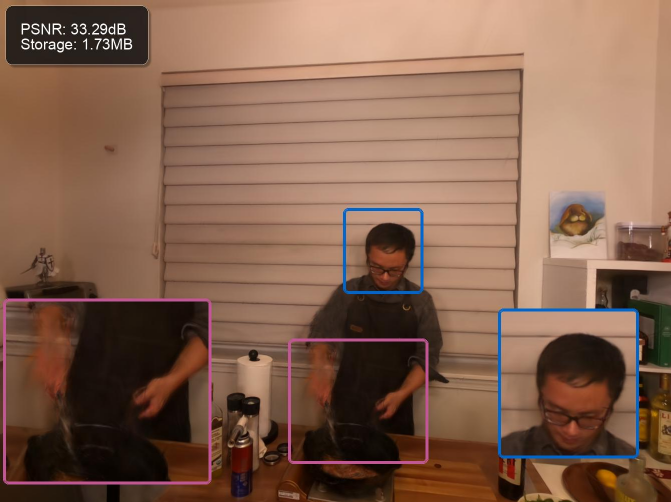}
\end{minipage}%
\begin{minipage}[t]{0.168\textwidth}
    \centering
    \parbox[c][0.42cm][c]{\linewidth}{\centering \small HiCoM}
    \includegraphics[width=\linewidth,keepaspectratio]{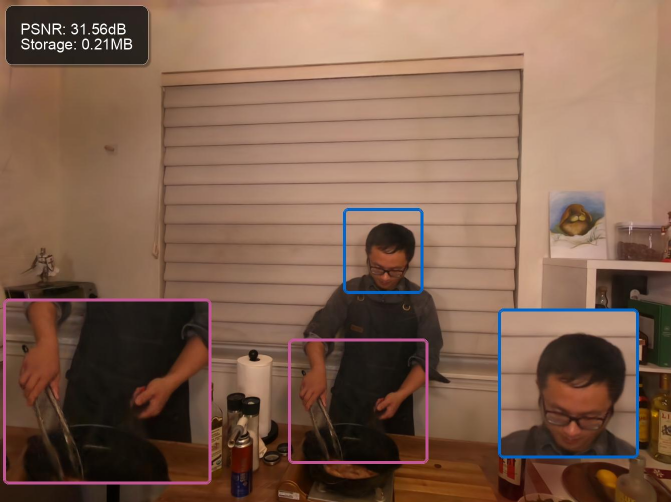}
\end{minipage}%
\begin{minipage}[t]{0.168\textwidth}
    \centering
    \parbox[c][0.42cm][c]{\linewidth}{\centering \small iFVC}
    \includegraphics[width=\linewidth,keepaspectratio]{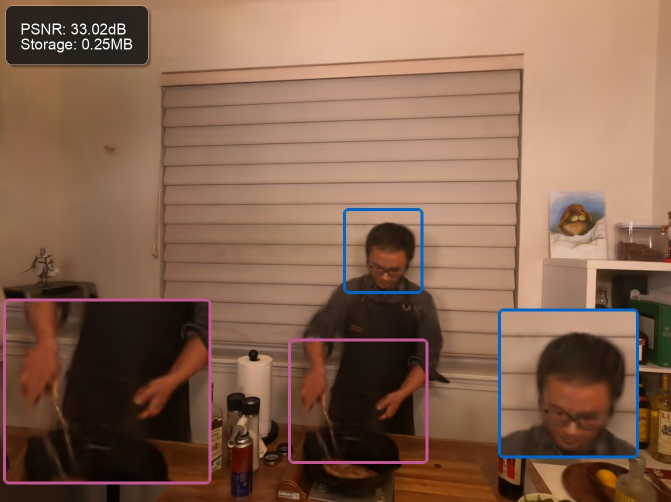}
\end{minipage}%
\begin{minipage}[t]{0.168\textwidth}
    \centering
    \parbox[c][0.42cm][c]{\linewidth}{\centering \small \textbf{SoLAR}}
    \includegraphics[width=\linewidth,keepaspectratio]{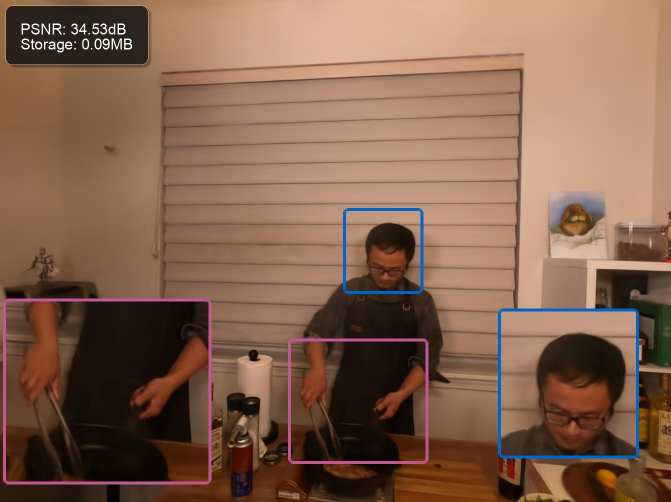}
\end{minipage}\\[-1pt]

\begin{minipage}[t]{0.168\textwidth}
    \centering
    \includegraphics[width=\linewidth,keepaspectratio]{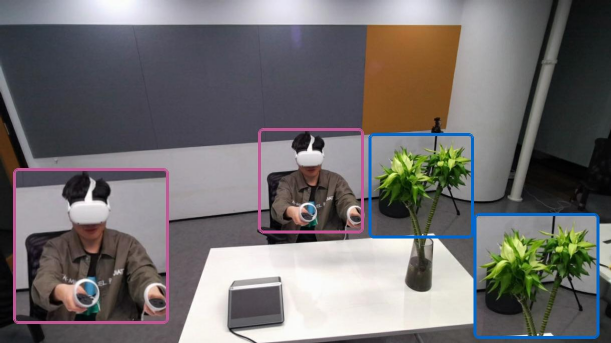}
\end{minipage}%
\begin{minipage}[t]{0.168\textwidth}
    \centering
    \includegraphics[width=\linewidth,keepaspectratio]{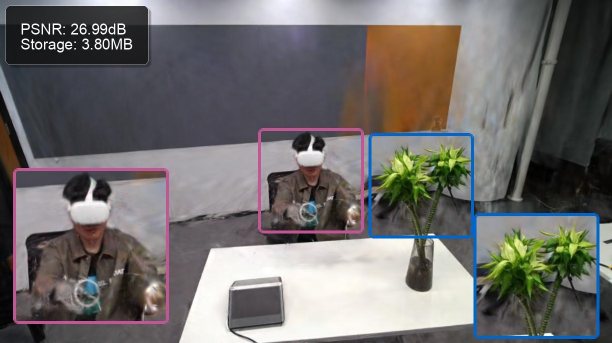}
\end{minipage}%
\begin{minipage}[t]{0.168\textwidth}
    \centering
    \includegraphics[width=\linewidth,keepaspectratio]{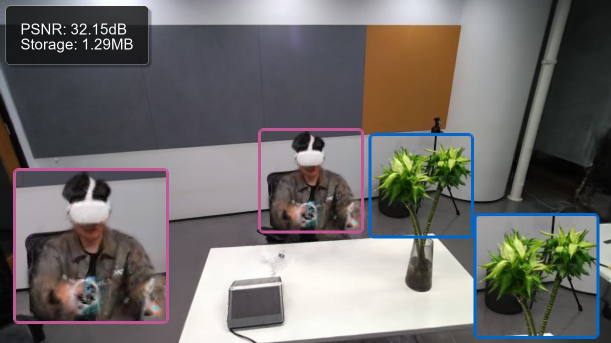}
\end{minipage}%
\begin{minipage}[t]{0.168\textwidth}
    \centering
    \includegraphics[width=\linewidth,keepaspectratio]{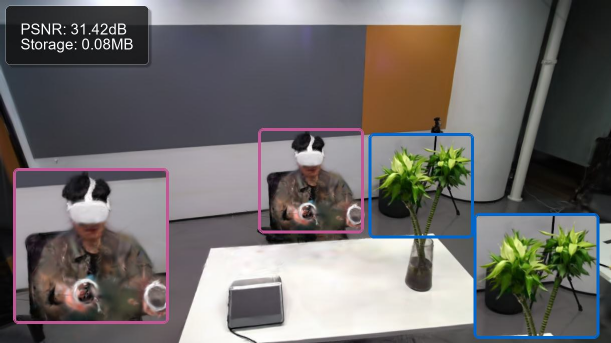}
\end{minipage}%
\begin{minipage}[t]{0.168\textwidth}
    \centering
    \includegraphics[width=\linewidth,keepaspectratio]{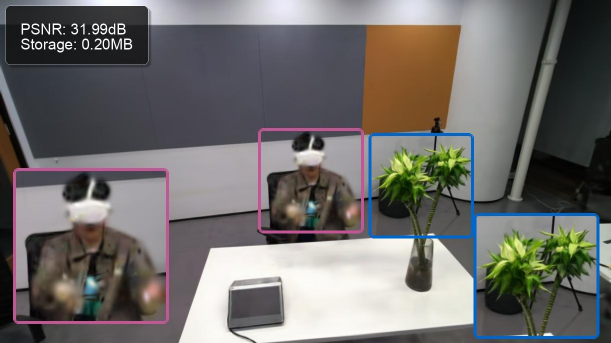}
\end{minipage}%
\begin{minipage}[t]{0.168\textwidth}
    \centering
    \includegraphics[width=\linewidth,keepaspectratio]{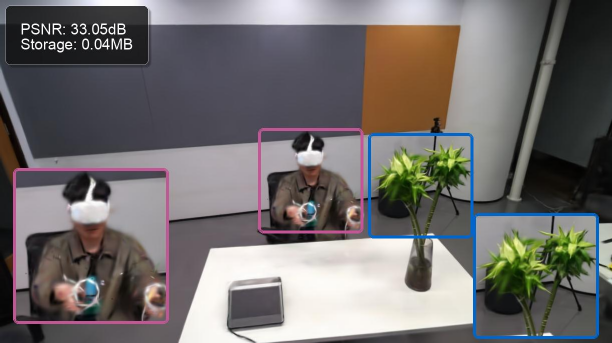}
\end{minipage}\\[-1pt]

\begin{minipage}[t]{0.168\textwidth}
    \centering
    \includegraphics[width=\linewidth,keepaspectratio]{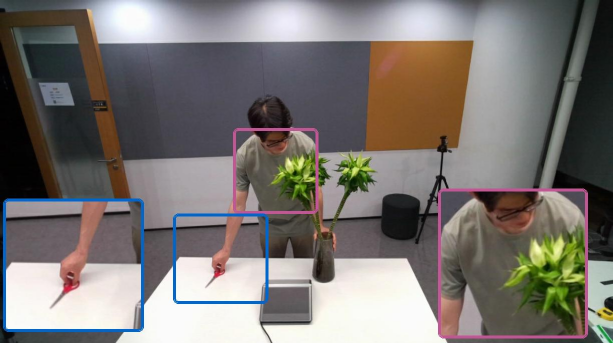}
\end{minipage}%
\begin{minipage}[t]{0.168\textwidth}
    \centering
    \includegraphics[width=\linewidth,keepaspectratio]{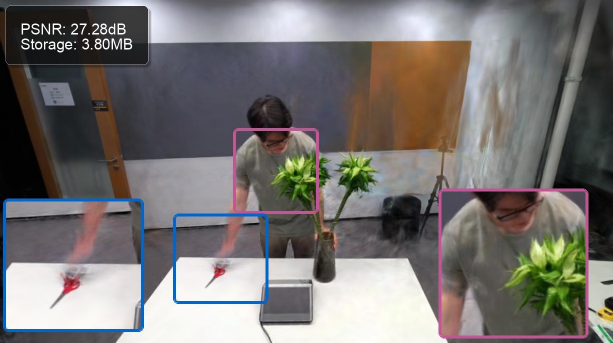}
\end{minipage}%
\begin{minipage}[t]{0.168\textwidth}
    \centering
    \includegraphics[width=\linewidth,keepaspectratio]{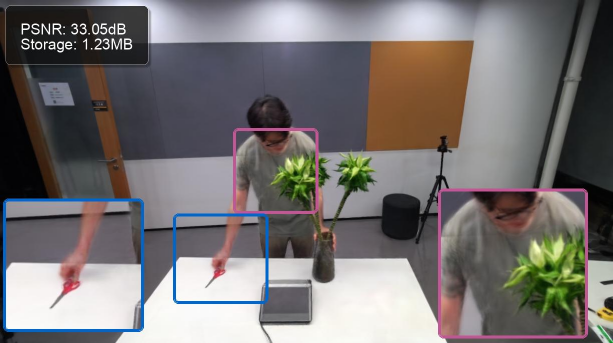}
\end{minipage}%
\begin{minipage}[t]{0.168\textwidth}
    \centering
    \includegraphics[width=\linewidth,keepaspectratio]{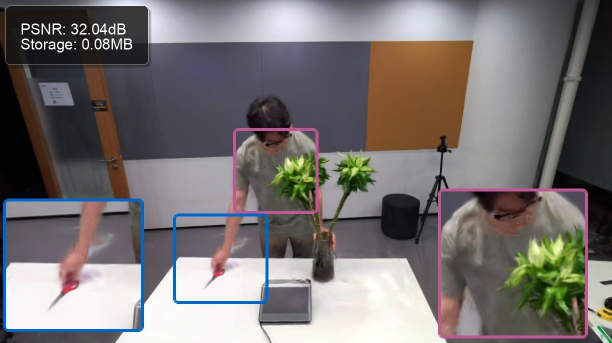}
\end{minipage}%
\begin{minipage}[t]{0.168\textwidth}
    \centering
    \includegraphics[width=\linewidth,keepaspectratio]{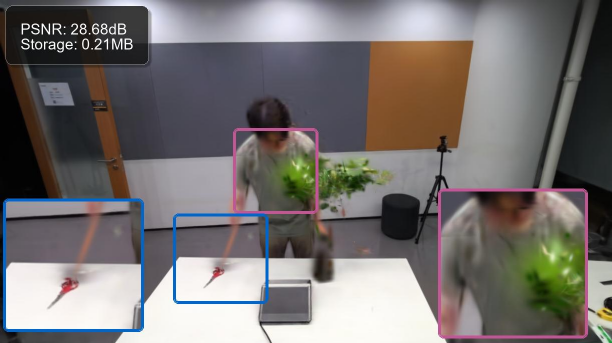}
\end{minipage}%
\begin{minipage}[t]{0.168\textwidth}
    \centering
    \includegraphics[width=\linewidth,keepaspectratio]{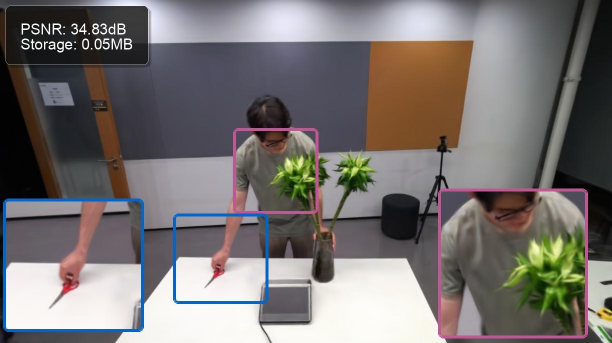}
\end{minipage}\\[-1pt]

\begin{minipage}[t]{0.168\textwidth}
    \centering
    \includegraphics[width=\linewidth,keepaspectratio]{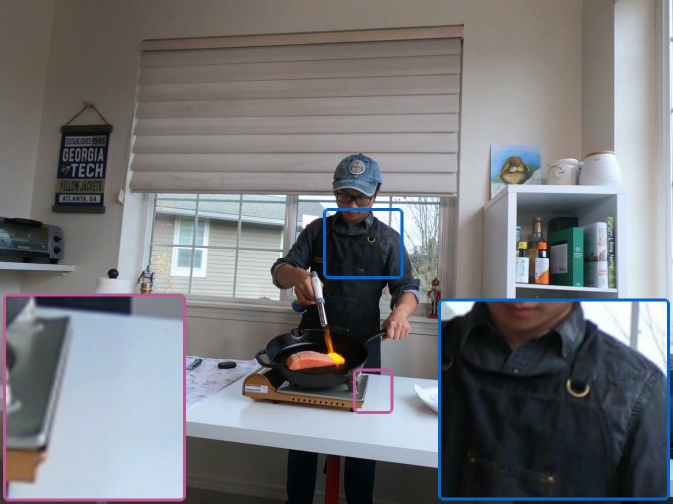}
\end{minipage}%
\begin{minipage}[t]{0.168\textwidth}
    \centering
    \includegraphics[width=\linewidth,keepaspectratio]{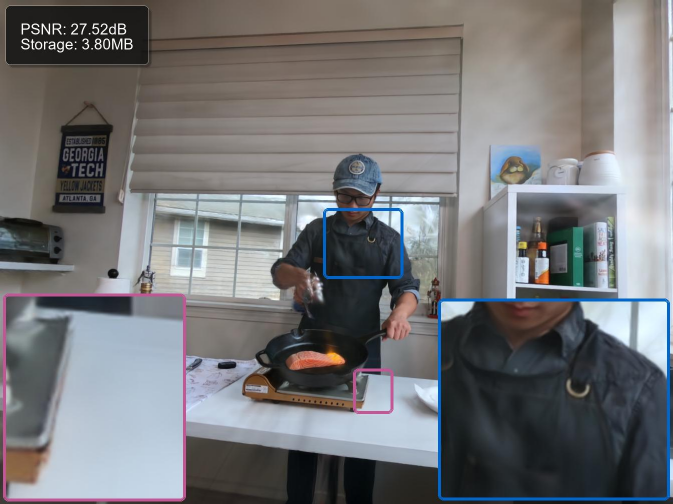}
\end{minipage}%
\begin{minipage}[t]{0.168\textwidth}
    \centering
    \includegraphics[width=\linewidth,keepaspectratio]{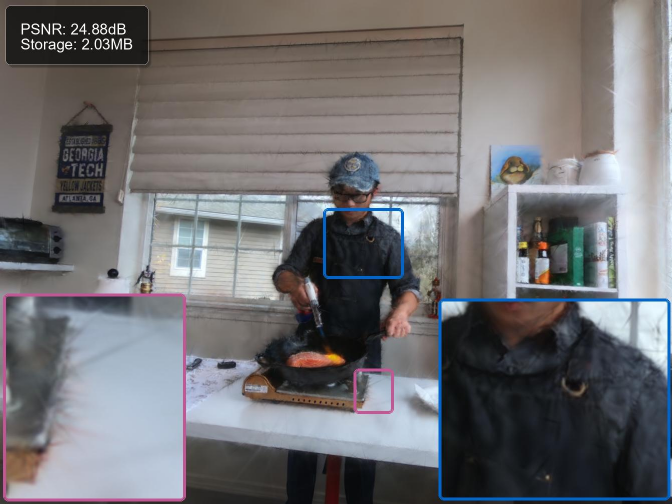}
\end{minipage}%
\begin{minipage}[t]{0.168\textwidth}
    \centering
    \includegraphics[width=\linewidth,keepaspectratio]{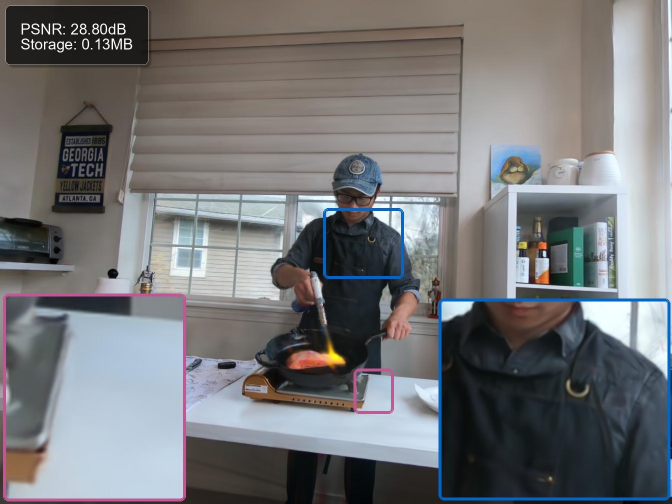}
\end{minipage}%
\begin{minipage}[t]{0.168\textwidth}
    \centering
    \includegraphics[width=\linewidth,keepaspectratio]{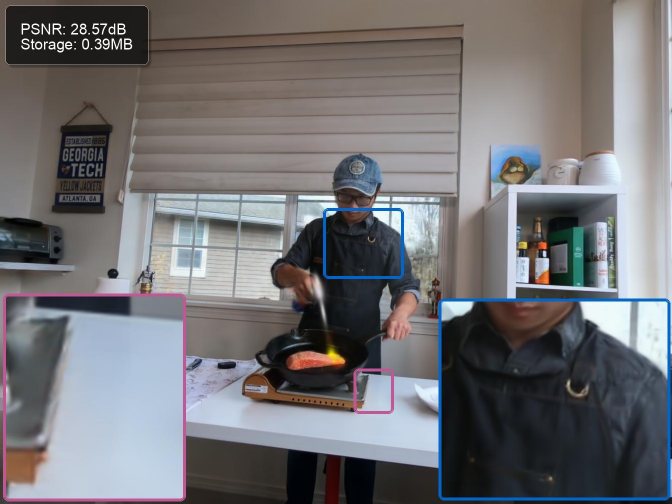}
\end{minipage}%
\begin{minipage}[t]{0.168\textwidth}
    \centering
    \includegraphics[width=\linewidth,keepaspectratio]{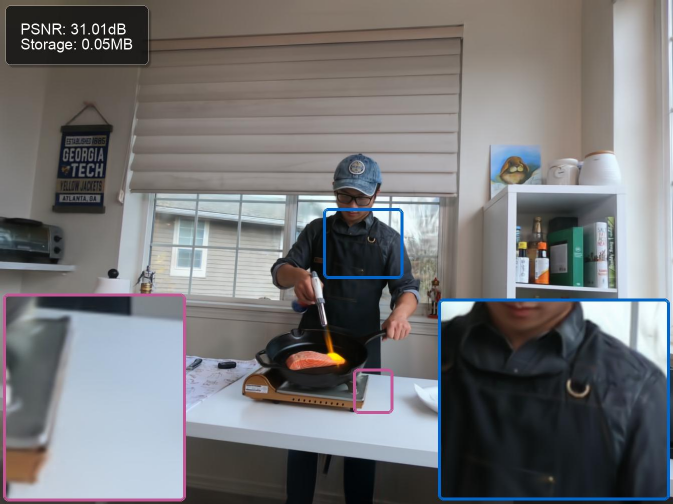}
\end{minipage}\\[-1pt]

\begin{minipage}[t]{0.168\textwidth}
    \centering
    \includegraphics[width=\linewidth,keepaspectratio]{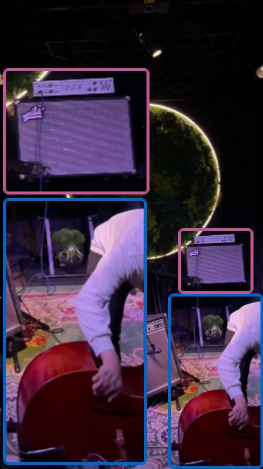}
\end{minipage}%
\begin{minipage}[t]{0.168\textwidth}
    \centering
    \includegraphics[width=\linewidth,keepaspectratio]{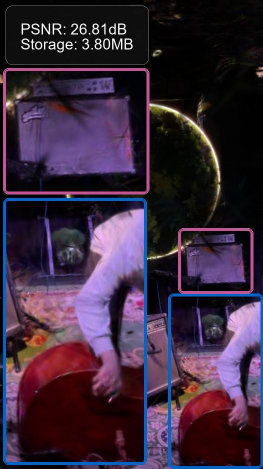}
\end{minipage}%
\begin{minipage}[t]{0.168\textwidth}
    \centering
    \includegraphics[width=\linewidth,keepaspectratio]{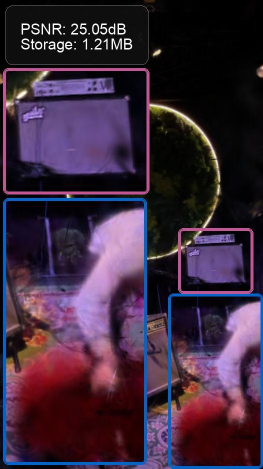}
\end{minipage}%
\begin{minipage}[t]{0.168\textwidth}
    \centering
    \includegraphics[width=\linewidth,keepaspectratio]{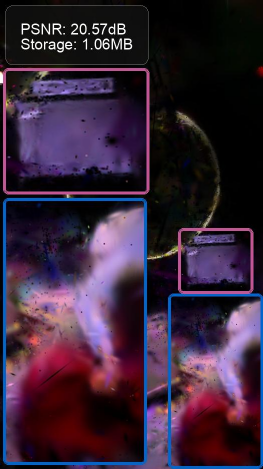}
\end{minipage}%
\begin{minipage}[t]{0.168\textwidth}
    \centering
    \includegraphics[width=\linewidth,keepaspectratio]{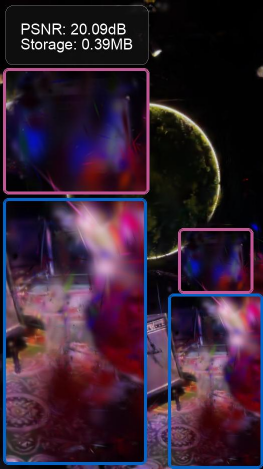}
\end{minipage}%
\begin{minipage}[t]{0.168\textwidth}
    \centering
    \includegraphics[width=\linewidth,keepaspectratio]{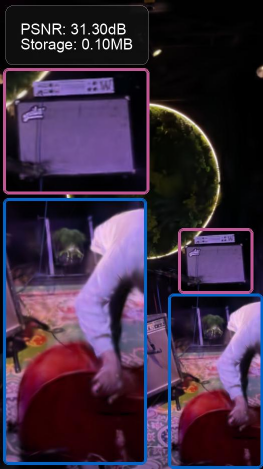}
\end{minipage}

\caption{
\textbf{Qualitative comparison.} \textbf{SoLAR} consistently reconstructs finer details and delivers the best visual quality across all evaluated datasets.
}
\label{fig:qualitative_results}
\end{figure*}
\begin{figure*}[t]
\centering
\setlength{\tabcolsep}{0pt}

\begin{minipage}[t]{0.168\textwidth}
    \centering
    \parbox[c][0.42cm][c]{\linewidth}{\centering \small Ground truth}
    \includegraphics[width=\linewidth,keepaspectratio]{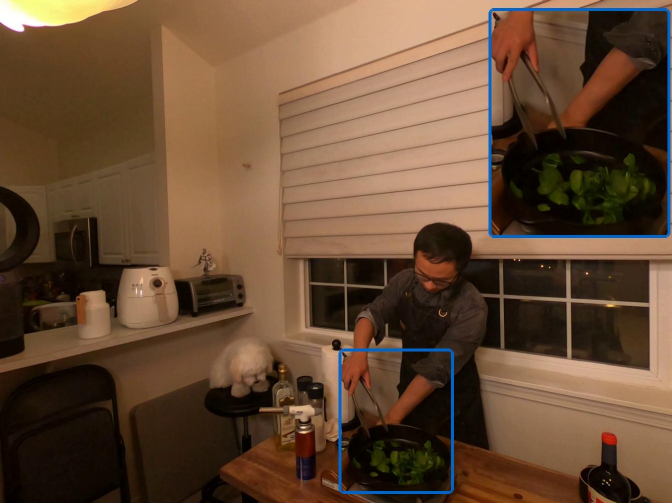}
\end{minipage}%
\begin{minipage}[t]{0.168\textwidth}
    \centering
    \parbox[c][0.42cm][c]{\linewidth}{\centering \small 3DGStream}
    \includegraphics[width=\linewidth,keepaspectratio]{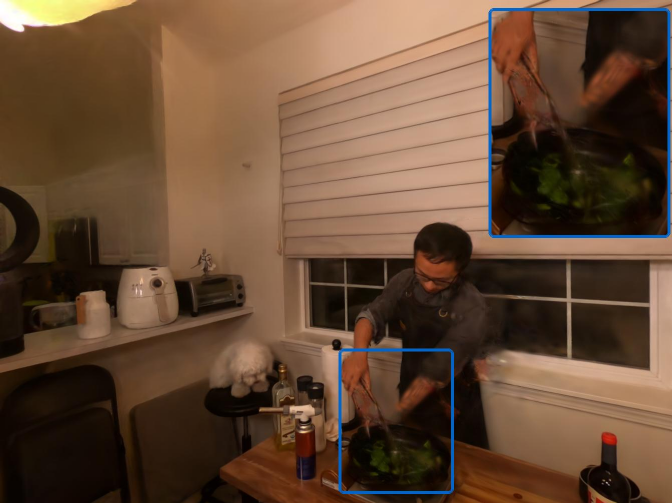}
\end{minipage}%
\begin{minipage}[t]{0.168\textwidth}
    \centering
    \parbox[c][0.42cm][c]{\linewidth}{\centering \small 4DGC}
    \includegraphics[width=\linewidth,keepaspectratio]{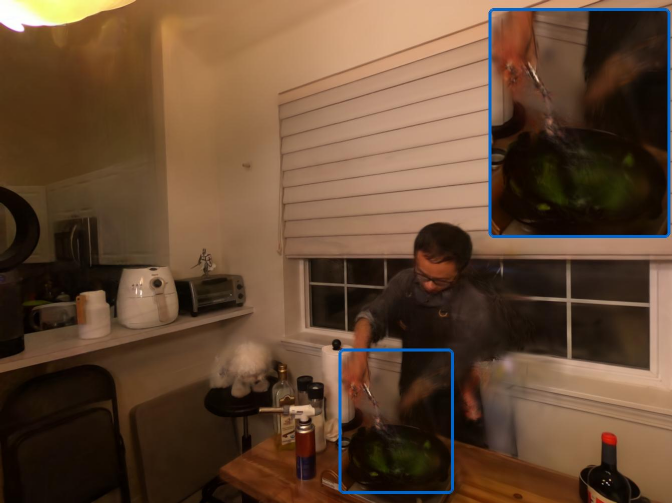}
\end{minipage}%
\begin{minipage}[t]{0.168\textwidth}
    \centering
    \parbox[c][0.42cm][c]{\linewidth}{\centering \small HiCoM}
    \includegraphics[width=\linewidth,keepaspectratio]{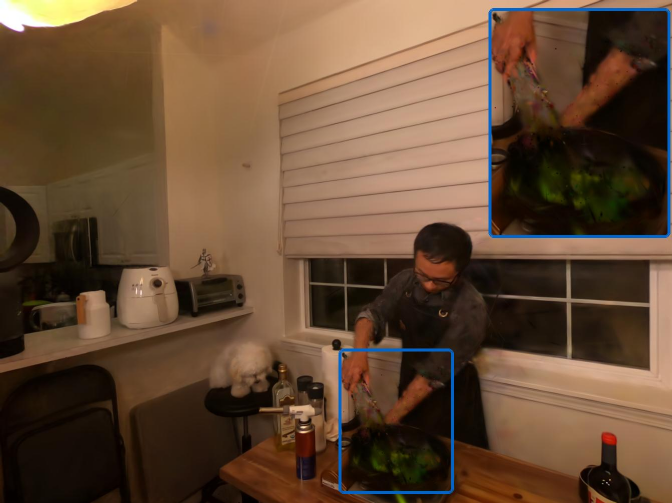}
\end{minipage}%
\begin{minipage}[t]{0.168\textwidth}
    \centering
    \parbox[c][0.42cm][c]{\linewidth}{\centering \small iFVC}
    \includegraphics[width=\linewidth,keepaspectratio]{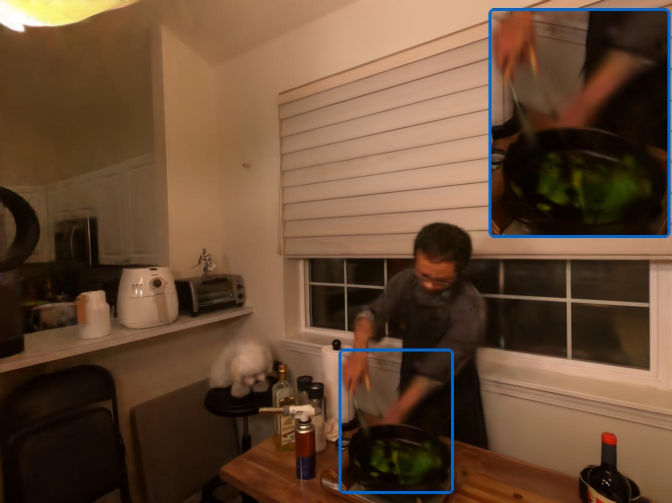}
\end{minipage}%
\begin{minipage}[t]{0.168\textwidth}
    \centering
    \parbox[c][0.42cm][c]{\linewidth}{\centering \small \textbf{SoLAR}}
    \includegraphics[width=\linewidth,keepaspectratio]{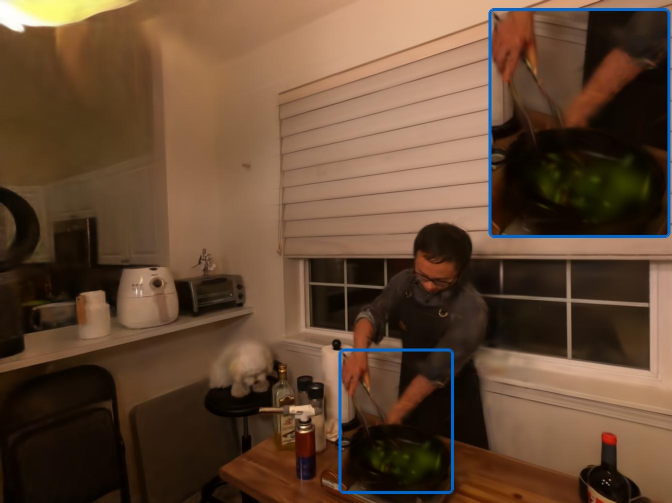}
\end{minipage}\\[-1pt]

\begin{minipage}[t]{0.168\textwidth}
    \centering
    \includegraphics[width=\linewidth,keepaspectratio]{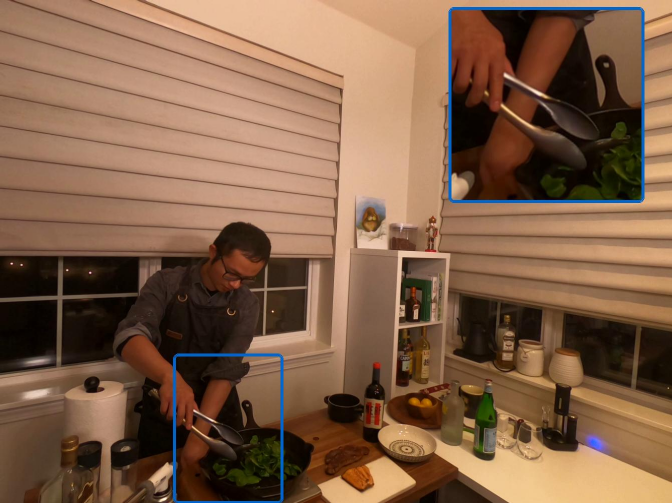}
\end{minipage}%
\begin{minipage}[t]{0.168\textwidth}
    \centering
    \includegraphics[width=\linewidth,keepaspectratio]{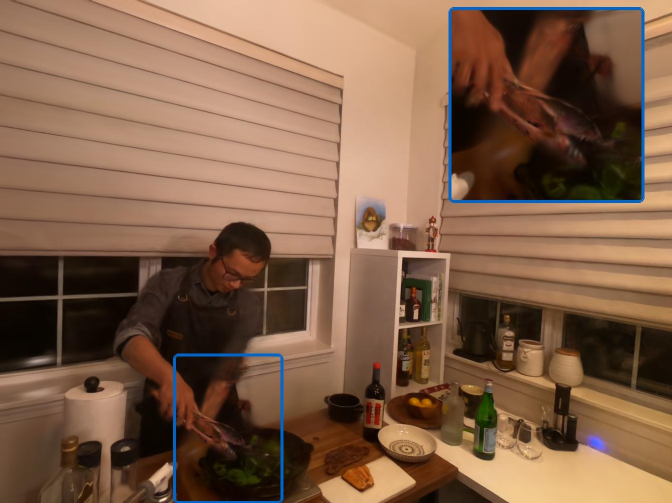}
\end{minipage}%
\begin{minipage}[t]{0.168\textwidth}
    \centering
    \includegraphics[width=\linewidth,keepaspectratio]{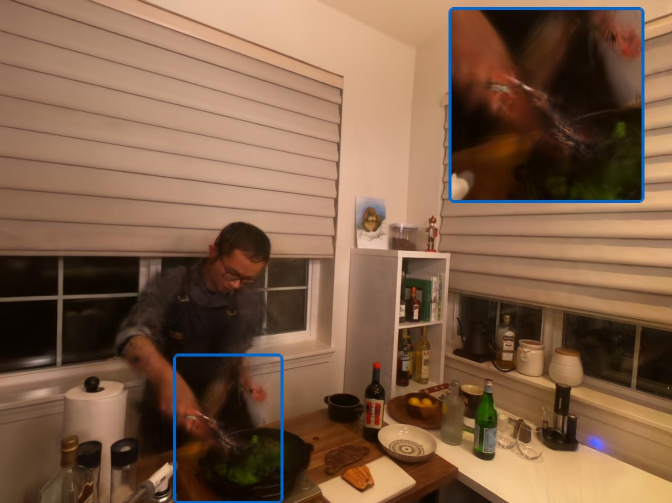}
\end{minipage}%
\begin{minipage}[t]{0.168\textwidth}
    \centering
    \includegraphics[width=\linewidth,keepaspectratio]{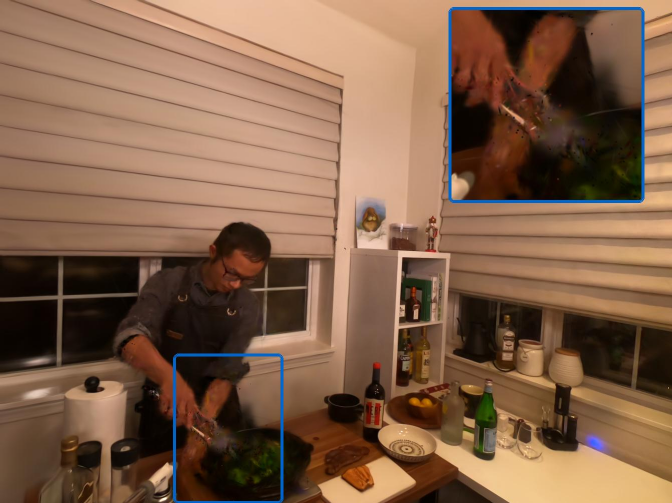}
\end{minipage}%
\begin{minipage}[t]{0.168\textwidth}
    \centering
    \includegraphics[width=\linewidth,keepaspectratio]{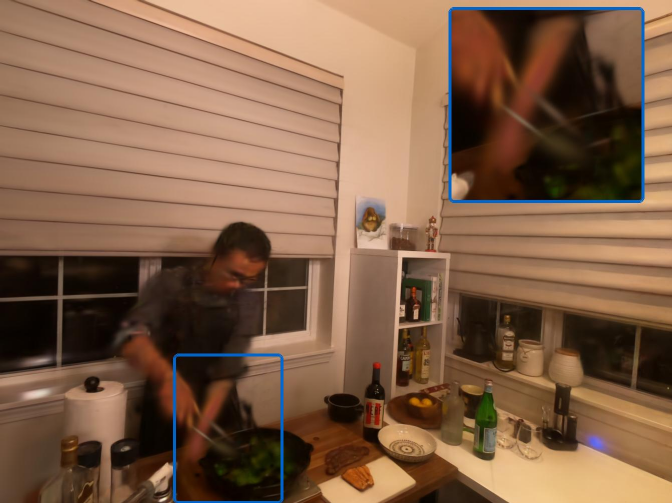}
\end{minipage}%
\begin{minipage}[t]{0.168\textwidth}
    \centering
    \includegraphics[width=\linewidth,keepaspectratio]{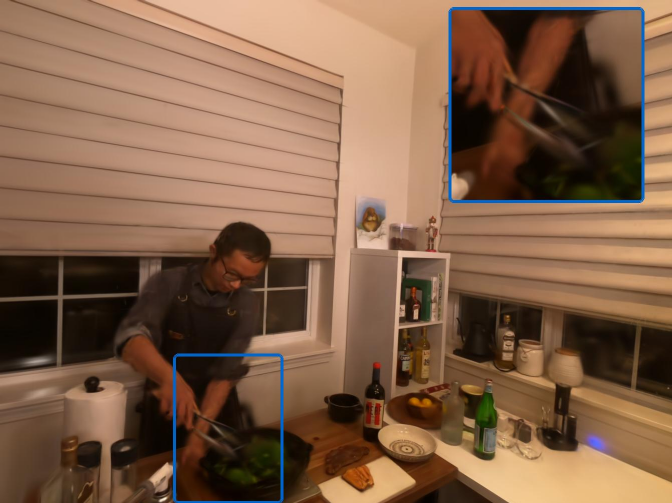}
\end{minipage}\\[-1pt]

\begin{minipage}[t]{0.168\textwidth}
    \centering
    \includegraphics[width=\linewidth,keepaspectratio]{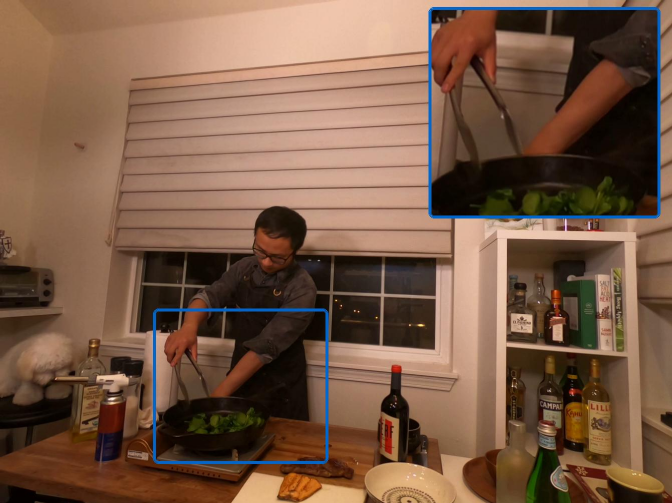}
\end{minipage}%
\begin{minipage}[t]{0.168\textwidth}
    \centering
    \includegraphics[width=\linewidth,keepaspectratio]{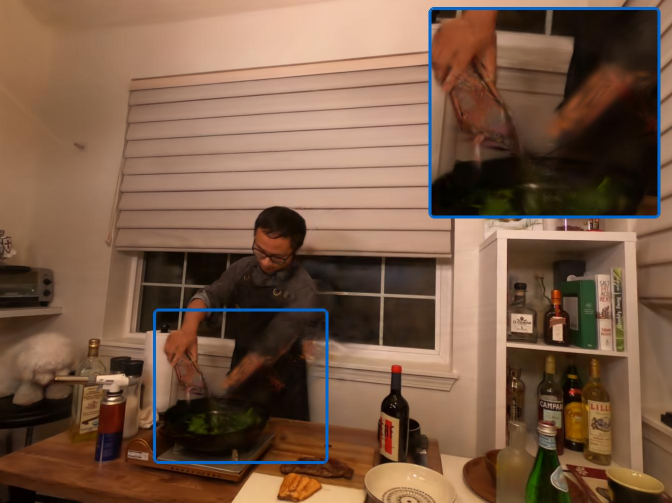}
\end{minipage}%
\begin{minipage}[t]{0.168\textwidth}
    \centering
    \includegraphics[width=\linewidth,keepaspectratio]{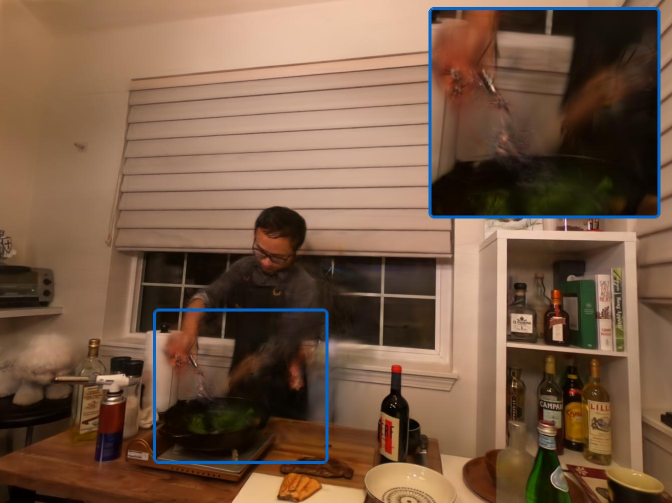}
\end{minipage}%
\begin{minipage}[t]{0.168\textwidth}
    \centering
    \includegraphics[width=\linewidth,keepaspectratio]{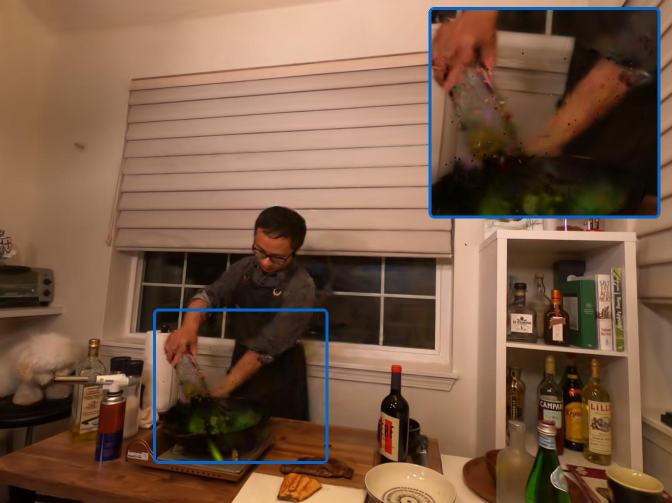}
\end{minipage}%
\begin{minipage}[t]{0.168\textwidth}
    \centering
    \includegraphics[width=\linewidth,keepaspectratio]{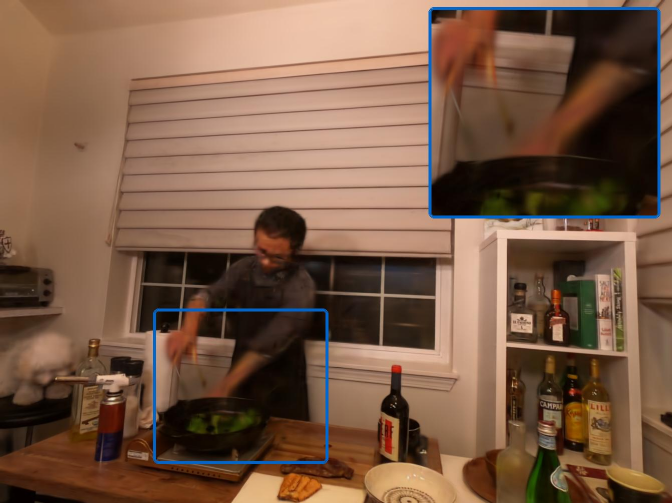}
\end{minipage}%
\begin{minipage}[t]{0.168\textwidth}
    \centering
    \includegraphics[width=\linewidth,keepaspectratio]{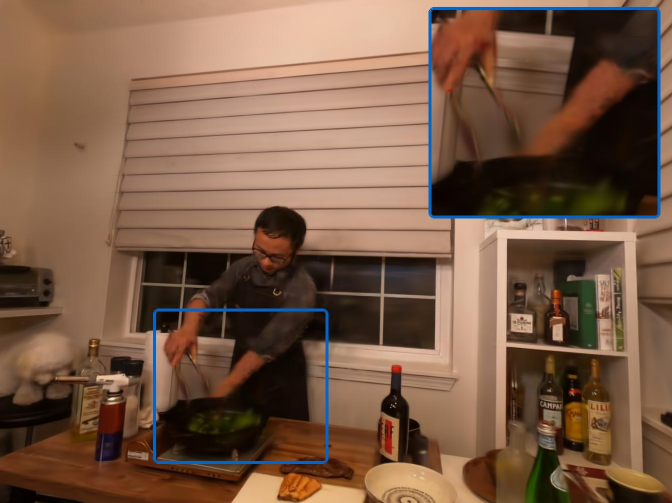}
\end{minipage}\\[-1pt]

\begin{minipage}[t]{0.168\textwidth}
    \centering
    \includegraphics[width=\linewidth,keepaspectratio]{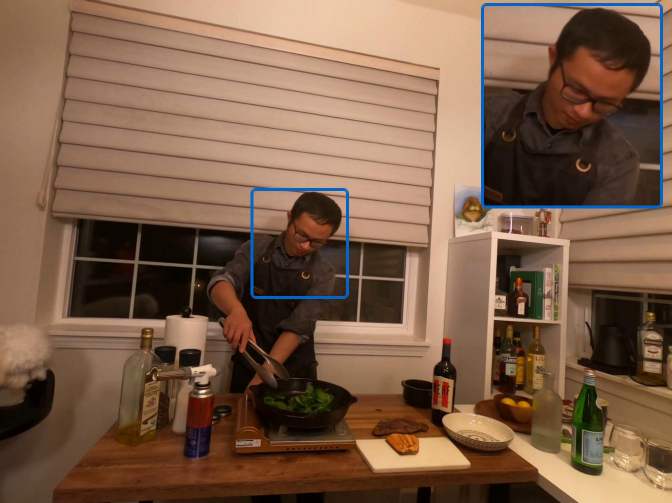}
\end{minipage}%
\begin{minipage}[t]{0.168\textwidth}
    \centering
    \includegraphics[width=\linewidth,keepaspectratio]{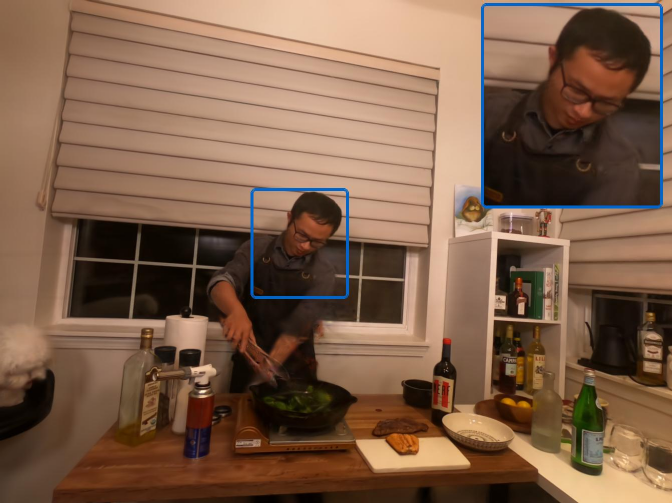}
\end{minipage}%
\begin{minipage}[t]{0.168\textwidth}
    \centering
    \includegraphics[width=\linewidth,keepaspectratio]{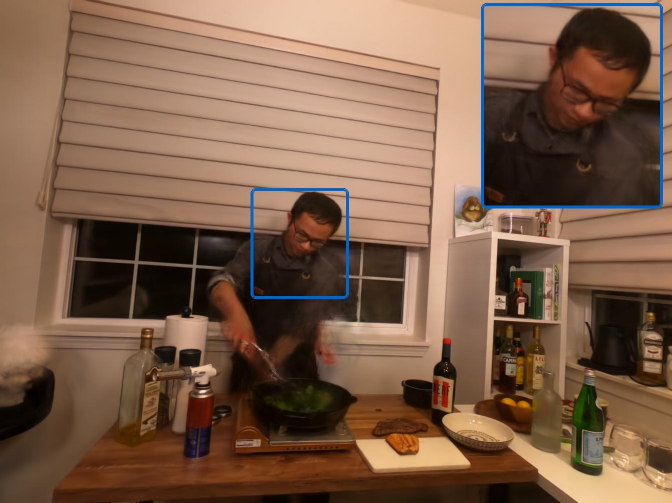}
\end{minipage}%
\begin{minipage}[t]{0.168\textwidth}
    \centering
    \includegraphics[width=\linewidth,keepaspectratio]{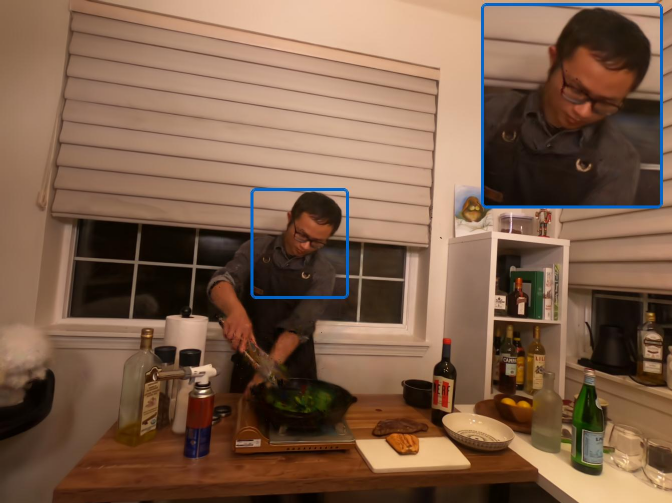}
\end{minipage}%
\begin{minipage}[t]{0.168\textwidth}
    \centering
    \includegraphics[width=\linewidth,keepaspectratio]{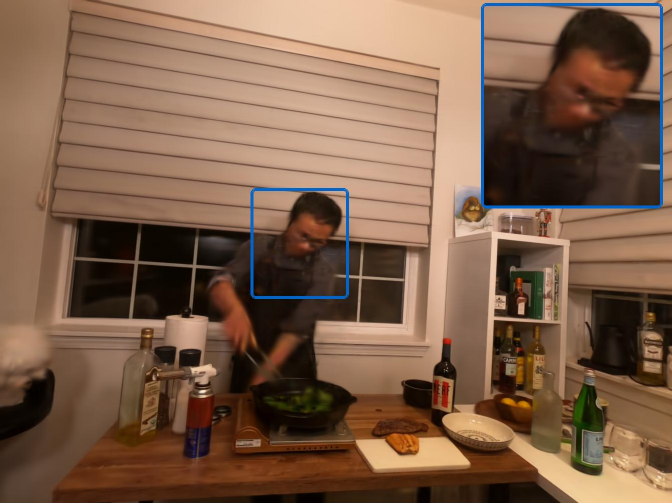}
\end{minipage}%
\begin{minipage}[t]{0.168\textwidth}
    \centering
    \includegraphics[width=\linewidth,keepaspectratio]{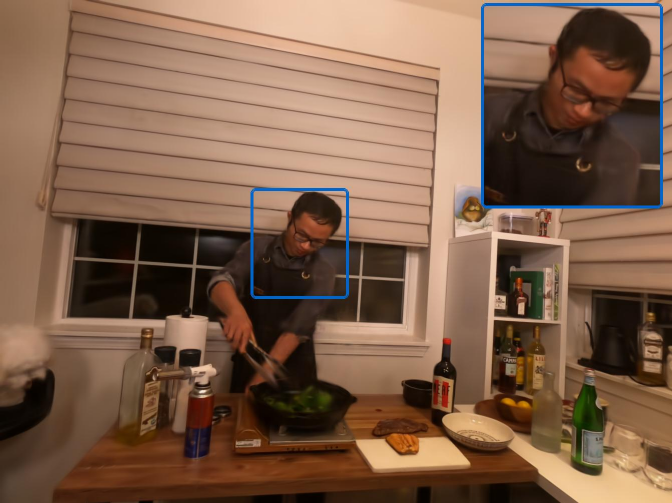}
\end{minipage}

\caption{
\textbf{Multi-view consistency comparison}. SoLAR delivers more stable and coherent renderings across novel viewpoints.
}
\label{fig:muiti_views_rendering_result}
\end{figure*}
\subsection{Experiment Results}
\label{sec:experiment:results}

\Paragraph{Quantitative Results.} Our proposed method \textbf{SoLAR} consistently achieves the best reconstruction quality while requiring the least storage across all evaluated datasets. Meanwhile, it maintains training and rendering speeds comparable to existing approaches, demonstrating a superior rate-distortion trade-off and remarkable computational efficiency. As shown in Tab.~\ref{tab:meetroom_n3dv}, Tab.~\ref{tab:bar_table}, Tab.~\ref{tab:quantitative_result_lbn1}, Tab.~\ref{tab:quantitative_result_lbn2}, and Tab.~\ref{tab:quantitative_result_zxc}, \textbf{SoLAR} consistently improves PSNR over recent state-of-the-art approaches by approximately \Embf{1}~dB on N3DV, \Embf{1.5}~dB on Meeting Room, \Embf{4}~dB on Bar, and \Embf{4}~dB on AvatarRex-lbn1. Notably, \textit{SoLAR-lite} further yields a dramatic reduction in storage overhead while still achieving the second-best reconstruction quality, surpassed only by \textbf{SoLAR}. Specifically, on the \textit{Meeting Room} dataset, \textbf{SoLAR} improves reconstruction fidelity by approximately \Embf{3}~dB while reducing storage requirements by \Embf{6}$\times$ relative to Recon-GS. Moreover, \textbf{SoLAR} and \textit{SoLAR-lite} achieve the top two PSNR and SSIM results, while reducing storage consumption by more than \Embf{200}$\times$ compared with 3DGStream. The superiority of the proposed method is further reflected on the long-horizon \textit{Bar} dataset, where \textbf{SoLAR} surpasses iFVC by approximately \Embf{4}~dB and simultaneously reduces storage overhead by \Embf{10}$\times$. Similarly, on the human-centric dataset \textit{AvatarRex-lbn1}, \textbf{SoLAR} is the only method that sustains a PSNR above \Embf{30}~dB, while maintaining an average storage overhead of merely \Embf{0.05}~MB. These results demonstrate that \textbf{SoLAR} exhibits superior cross-dataset generalization ability and maintains favorable robustness under long-horizon streaming scenarios. In particular, the proposed framework effectively mitigates error accumulation in LFVV and maintains stable performance over extended sequences. As shown in Fig.~\ref{fig:teaser}, competing methods suffer a clear quality drop after 300--500 frames, whereas \textbf{SoLAR} remains stable throughout the sequence. Even when mild early-stage degradation occurs, \textbf{SoLAR} recovers and suppresses subsequent error propagation, preventing gradual performance decline. This stability comes from the complementary effects of AAD and LaDAR. Specifically, AAD improves the modeling of non-rigid transformations, enhancing robustness against errors introduced by temporal propagation. Meanwhile, LaDAR identifies and corrects discrepancies in latent representations, effectively limiting their continuous accumulation across frames. Together, these modules mitigate error accumulation without relying on GOP partitioning, resulting in consistently excellent performance across diverse datasets under the considered setting.

\Paragraph{Qualitative Results.} Qualitative results further corroborate the effectiveness of \textbf{SoLAR}. As shown in Fig.~\ref{fig:teaser}, on the \textit{Bar} dataset, as the sequence progresses and the object undergoes rapid, large-amplitude motion, the competing method iFVC exhibits noticeable rendering degradation, including increasingly blurred outputs, loss of fine object details, and progressive deterioration in visual fidelity. By contrast, \textbf{SoLAR} maintains stable reconstruction quality throughout the sequence. This performance gap becomes even more striking when the object is temporarily occluded or disappears from view. Baseline methods exhibit noticeable failure cases, where the rendered appearance collapses and the object becomes difficult to distinguish, while \textbf{SoLAR} still preserves faithful visual details and superior appearance consistency. Fig.~\ref{fig:qualitative_results} further presents qualitative comparisons between \textbf{SoLAR} and representative baselines, including 3DGStream, 4DGC, HiCoM, and iFVC, across five scenes: \textit{Sear Steak}, \textit{Vrheadset}, \textit{Trimming}, \textit{Flame Salmon}, and \textit{Bar}. Across diverse motion patterns and scene dynamics, competing methods struggle to preserve clear local structures and stable appearance, especially in regions with pronounced non-rigid motion. This is mainly caused by limited capacity for complex motion modeling and accumulated errors during long-horizon temporal propagation, which progressively degrade later-frame rendering. In contrast, \textbf{SoLAR} addresses these limitations through the complementary effects of AAD and LaDAR, preserving finer details, clearer structures, and more stable appearance throughout the sequence.
This advantage is prominently observed in the final frame of \textit{Bar} (final row), where the person places the instrument down. Under this challenging motion scenario, all baseline methods produce severely blurred renderings with distorted geometric structures and substantial loss of fine details; some methods even suffer from complete rendering failure and unnatural visual artifacts. By contrast, \textbf{SoLAR} achieves higher visual fidelity and more faithful recovery of local structures, remaining consistently closer to the ground truth than all competing approaches. 
Taken together, these qualitative comparisons show that \textbf{SoLAR} provides more faithful long-horizon reconstruction, with stronger detail preservation, higher visual fidelity, and more stable temporal behavior throughout the sequence.

\Paragraph{Multi-view consistency.} To evaluate the robustness of multi-view consistency, we further conduct qualitative comparisons under varying camera viewpoints. As shown in Fig.~\ref{fig:muiti_views_rendering_result}, the \textit{Cook Spinach} scene contains challenging contact-intensive regions, where baseline methods often fail to maintain stable geometry and appearance across viewpoints. In the enlarged comparisons, they produce blurred contours and unstable contact geometry between the hand and tongs, over-smoothed and view-dependent distortions for the vegetables, and unnatural changes in the relative position between the tongs and wok rim. These failure cases primarily arise from the limited capacity of the competing methods to model dynamic scenes involving fine-grained non-rigid contact. By contrast, \textbf{SoLAR} effectively alleviates these limitations. In the enlarged views, it preserves clearer local geometry, more coherent motion patterns, and more stable cross-view spatial organization in contact-intensive regions. In particular, the relative position between the tongs and the wok rim remains consistent across viewpoints. The grasping configuration between the hand and the tongs is preserved more coherently. Meanwhile, the vegetables inside the wok retain a more stable distribution and appearance. Overall, \textbf{SoLAR} yields cleaner and more coherent novel-view renderings, better maintains the structural integrity of local interactions, and demonstrates excellent multi-view consistency for free-viewpoint video synthesis under unconstrained camera motion and continuous view transitions.

\begin{table}[t]
\centering
\caption{\textbf{Ablation study of key modules in SoLAR.} Method with $^{\dag}$ enables GOP partitioning.}
\label{tab:ablation_module}

\scriptsize
\setlength{\tabcolsep}{2pt}
\renewcommand{\arraystretch}{0.95}

\begin{tabularx}{\linewidth}{l *{5}{C}}
\toprule
Method 
& PSNR$\uparrow$ 
& LPIPS$\downarrow$ 
& Storage$\downarrow$ 
& Train$\downarrow$ 
& Render$\uparrow$ \\
& (dB) & & (MB) & (min) & (FPS) \\
\midrule
w/o AAD & \cellsecond 32.34 & \cellsecond 0.147 & 0.07 & \cellsecond 0.826 & \cellbest{231} \\
w/o LaDAR & 32.22 & \cellsecond 0.147 & \cellsecond 0.06 & 0.851 & 176 \\
w/o AAD \& LaDAR & 31.92 & 0.151 & 0.20 & \cellbest{0.820} & \cellsecond 206 \\
w/o AAD \& LaDAR$^{\dag}$ & 32.03 & 0.150 & 0.20 & 0.834 & 205 \\
SoLAR~(full) & \cellbest 32.45 & \cellbest 0.145 & \cellbest 0.05 & 0.852 & 180 \\
\bottomrule
\end{tabularx}
\end{table}
\begin{figure}[t]
  \centering
  \includegraphics[
    width=1\columnwidth,
    trim=0 7 0 0,
    clip
  ]{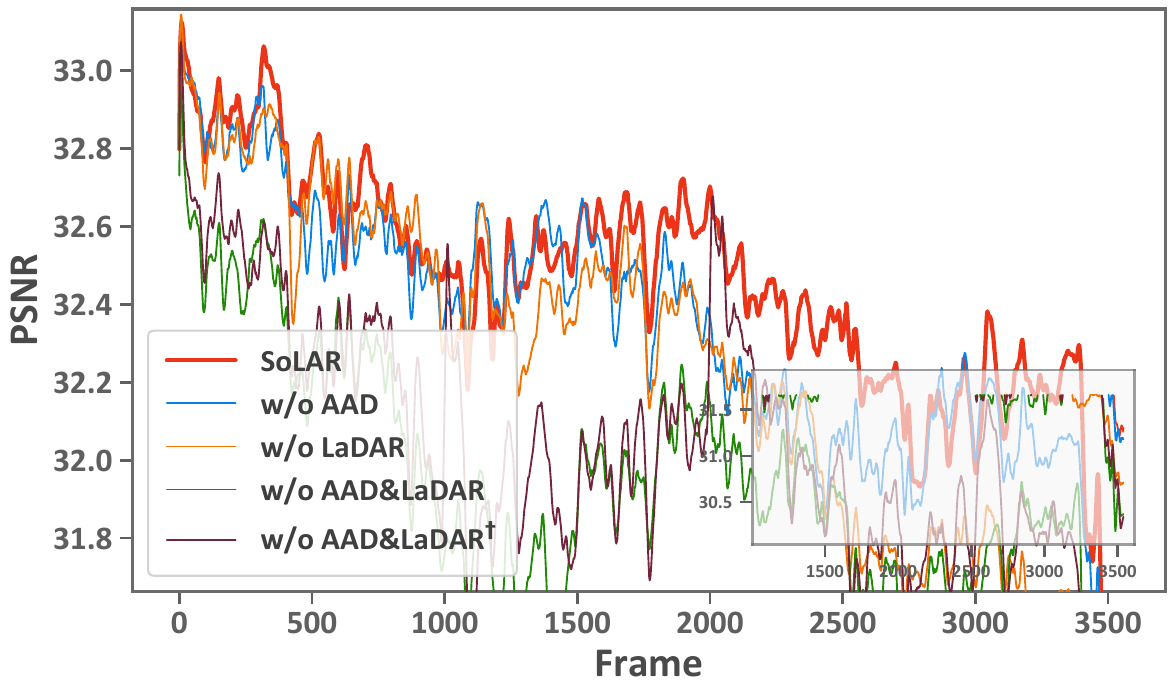}
  \caption{
    \textbf{Reconstruction quality trends across SoLAR ablation variants.} 
    Variants incorporating GOP partitioning are marked with $^{\dag}$.
  } 
  \label{Fig:ablation_trend}
\end{figure}
\subsection{Ablation Study}
\label{sec:experiment:ablation}
Since our work targets LFVV reconstruction, we conduct the ablation study on the challenging long-horizon \emph{Bar} dataset to evaluate the individual contributions of AAD and LaDAR. \emph{Bar} contains more than 3{,}000 frames and exhibits large-amplitude motion, temporary disappearance of scene content, and substantial long-term temporal variations. These characteristics make it a representative setting in which temporal drift and accumulated errors become major sources of degradation. As summarized in Tab.~\ref{tab:ablation_module} and Fig.~\ref{Fig:ablation_trend}, both modules improve reconstruction quality and reduce storage consumption, while introducing only negligible additional training cost.

\Paragraph{Effectiveness of AAD.} AAD improves the overall performance of \textbf{SoLAR} by enhancing the modeling capacity of dynamic anchors for non-rigid transformations. Specifically, it employs a soft masking mechanism to dynamically activate informative anchors and suppress less relevant ones. Unlike permanent pruning, anchors suppressed by the mask at the current timestamp are not removed from the representation and can be reactivated in subsequent frames. This temporal reversibility provides a flexible and adaptive representation, enabling the model to accommodate evolving scene content, including newly appearing objects and complex local motion, without introducing additional anchors. The quantitative results in Tab.~\ref{tab:ablation_module} confirm the effectiveness of this design. Compared with the variant without both AAD and LaDAR, introducing AAD improves PSNR by approximately 0.3~dB. More importantly, the storage overhead is reduced from 0.20~MB to 0.06~MB, corresponding to a more than $3\times$ reduction. These results indicate that AAD not only improves reconstruction fidelity, but also leads to a more compact BTC representation, demonstrating its effectiveness in improving both scene compactness and representational capacity.

\Paragraph{Effectiveness of LaDAR.} LaDAR is designed to recalibrate the Gaussian attribute network $\mathbf{N}_G$ while keeping the anchor features fixed. Accordingly, LaDAR updates the mapping from anchor latent features to Gaussian attributes by fine-tuning $\mathbf{N}_G$ without altering the anchor representation itself. This mechanism enables the decoded Gaussian attributes to remain better aligned with the current latent representation, thereby suppressing temporal discrepancies that would otherwise propagate to subsequent frames. As shown in Tab.~\ref{tab:ablation_module}, compared with the variant without both AAD and LaDAR, introducing LaDAR improves reconstruction quality by approximately 0.4~dB, while reducing the storage overhead from 0.20~MB to 0.07~MB, demonstrating that LaDAR suppresses temporal representation drift and preserves compactness. Moreover, removing LaDAR results in a noticeable drop in rendering speed, indicating that aligning the decoded Gaussian attributes with the current latent representation can further accelerate rendering. Overall, these results demonstrate that the introduction of LaDAR delivers superior performance and effectively mitigates error propagation, while sustaining compact storage overhead and practical rendering efficiency.

\Paragraph{Detailed Comparison with iFVC.} We observe that even after disabling the AAD and LaDAR (denoted as w/o AAD \& LaDAR), the proposed framework still outperforms iFVC on the long-horizon Bar dataset. This superiority stems from their different dynamic-anchor formulations. iFVC keeps anchor positions fixed during inter-frame transformations and only updates anchor features, whereas \textbf{SoLAR} jointly transforms both anchor positions and features across frames. This joint transformation provides \textbf{SoLAR} with a more flexible temporal representation, enabling it to better capture scene variations and improve robustness in long-sequence video reconstruction.

\Paragraph{Storage Analysis.} To provide a more fine-grained analysis of the model storage cost, we examine the module-wise overhead of AAD and LaDAR and their impact on representation compactness.
Instead of assigning separate learnable parameters to each anchor, AAD uses a compact anchor mask network $\mathbf{N}_m$ to implicitly encode activation patterns, introducing only about $0.0026$~MB overhead per frame. When severe discrepancies arise between the anchor latent features $f$ and the Gaussian attribute network $\mathbf{N}_G$, the correspondence $\mathcal{K}$ encoded in $\mathbf{N}_G$ is recalibrated. The updated $\mathbf{N}_G$ is then stored and propagated to subsequent frames. Benefiting from the compact architecture of $\mathbf{N}_G$, this recalibration process introduces only limited storage and transmission overhead, requiring approximately $0.03$~MB. In addition, extensive experiments indicate a clear storage characteristic: the storage cost tends to increase in more challenging temporal reconstruction scenarios, where more complex motion and appearance variations must be modeled. For example, in the last 200 frames of \textit{Bar}, the storage cost increases from $0.05$~MB to $0.18$~MB per frame even with LaDAR enabled. This behavior can be attributed to the BTC network, which is responsible for modeling inter-frame feature transformations. Under severe motion variations, the BTC network needs to encode more intricate temporal transformation patterns, resulting in more complex network parameters and substantially higher storage consumption.

\Paragraph{Rendering Speed.} In terms of rendering efficiency, although the full \textbf{SoLAR} does not achieve the highest rendering speed, it remains competitive with existing methods, as shown in Tab.~\ref{tab:meetroom_n3dv}. This trade-off mainly comes from the anchor-based representation, which requires an additional decoding step to derive Gaussian attributes from anchors, unlike explicit Gaussian methods that directly rasterize available primitives. Besides, AAD incurs additional cost during rendering because dynamic suppression requires evaluating the anchor mask network $\mathbf{N}_m$.
Nevertheless, \textbf{SoLAR} maintains favorable efficiency partly due to LaDAR. By correcting latent discrepancies, LaDAR improves the alignment between the Gaussian attribute network $\mathbf{N}_G$ and time-varying anchor features, making Gaussian attribute decoding more stable and efficient. It also reduces the burden on BTC, allowing it to focus on actual temporal transformations rather than accumulated discrepancies. As a result, the model can capture frame-to-frame variations more efficiently, which facilitates Gaussian attribute decoding and contributes to improved rendering speed during training.

\Paragraph{Training Speed.} As reported in Tab.~\ref{tab:bar_table}, the proposed method achieves the shortest training time among recent state-of-the-art approaches. To further analyze the training cost, we decompose the runtime of different components in Tab.~\ref{tab:ablation_module}. Although AAD and LaDAR introduce additional computation, their overhead remains marginal even on the long-sequence \textit{Bar} dataset. Specifically, AAD introduces extra mask computation and may reduce rendering FPS, but it improves compactness and reconstruction quality by suppressing less relevant anchors, adding only 0.03~minutes of training time compared with the variant without both AAD and LaDAR. LaDAR also introduces only about 0.006~minutes of overhead, since recalibration is triggered infrequently over the sequence.

\Paragraph{GOP Partitioning.}  
Notably, GOP partitioning brings only limited performance gains. As shown in Tab.~\ref{tab:bar_table} and Tab.~\ref{tab:ablation_module}, applying GOP partitioning to iFVC and to the variant w/o AAD \& LaDAR improves PSNR by only about 0.1~dB, while introducing considerable latency overhead. Although periodic resets can partially alleviate long-range error accumulation, they also disrupt inter-frame temporal consistency and introduce additional computational delay, thereby reducing real-time efficiency. In contrast, \textbf{SoLAR} mitigates error propagation through AAD and LaDAR within a unified streaming framework, avoiding explicit sequence partitioning while preserving both temporal consistency and overall efficiency.
\begin{figure}[t]
  \centering

  \includegraphics[
    width=0.92\linewidth,
    height=0.16\textheight,
    keepaspectratio,
    trim=8 6 8 6,
    clip
  ]{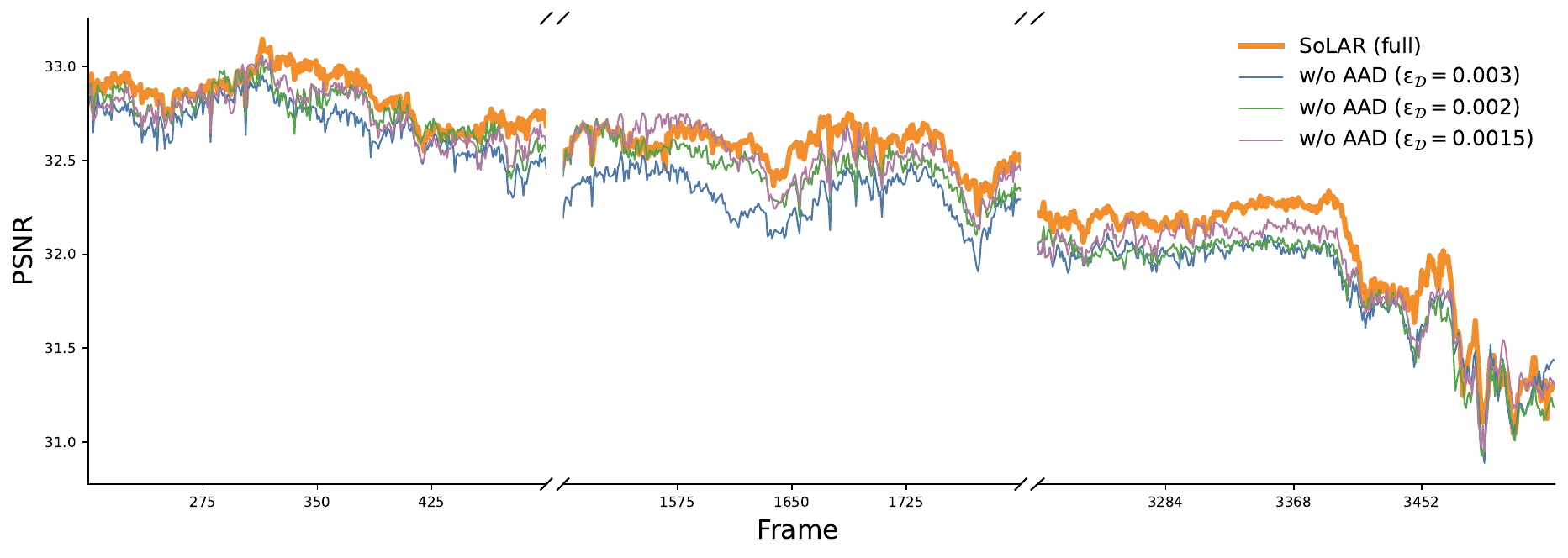}
  \centerline{\scriptsize (a) Reconstruction quality}

  \vspace{2pt}

  \includegraphics[
    width=0.92\linewidth,
    height=0.16\textheight,
    keepaspectratio,
    trim=8 6 8 6,
    clip
  ]{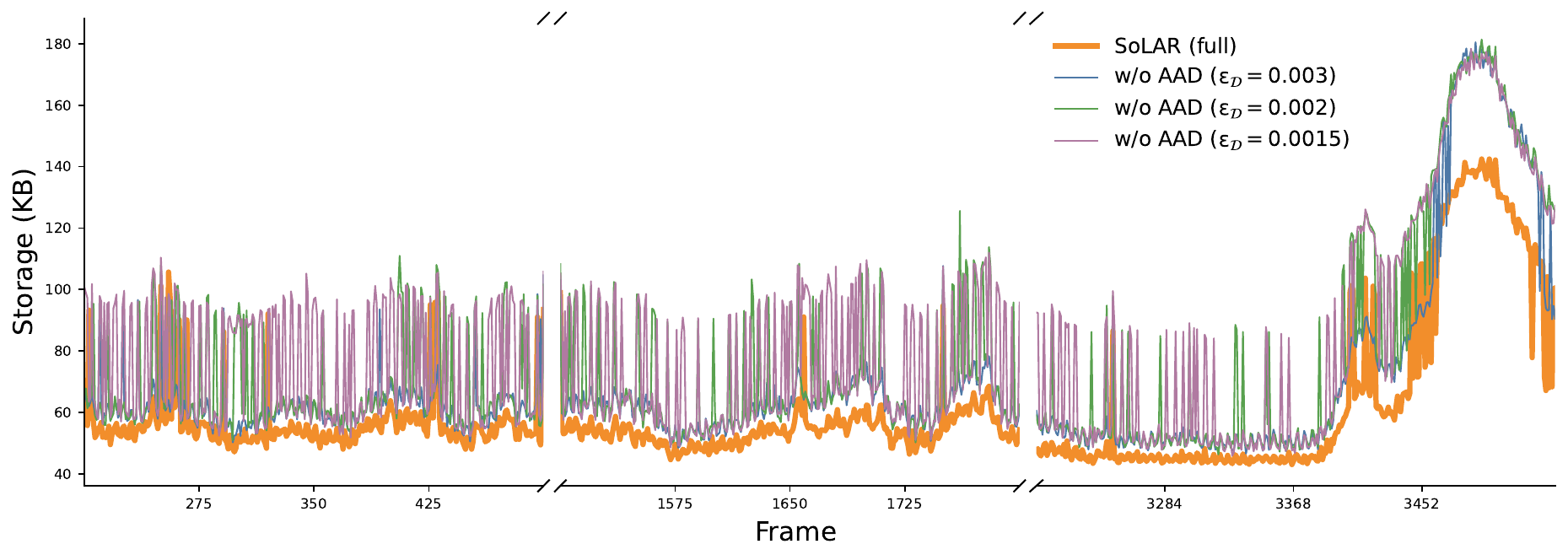}
  \centerline{\scriptsize (b) Storage overhead}

  \vspace{-4pt}

  \caption{\textbf{Temporal Dynamics of Reconstruction Quality and Storage Overhead.}
  Adjusting recalibration threshold controls the recalibration frequency.
  Top: temporal reconstruction quality. Bottom: temporal storage overhead.}
  \label{fig:psnr_storage_curve}
\end{figure}
\begin{table*}[t]
    \caption{\textbf{Examination of key hyperparameters in AAD and LaDAR.}}
    \label{tab:ablation_key_hyperparameters}
    \centering
    \footnotesize
    \setlength{\tabcolsep}{4pt}
    \renewcommand{\arraystretch}{0.95}

    \begin{tabularx}{0.92\textwidth}{lccCCCC}
        \toprule
        Module & Hyperparameter & Value
        & PSNR(dB)$\uparrow$
        & SSIM$\uparrow$
        & Storage(MB)$\downarrow$
        & Render(FPS)$\uparrow$ \\
        \midrule

        \multirow{7}{*}{AAD}
        & \multirow{4}{*}{\centering $\lambda_s$}
        & 0.1
        & 31.91
        & 0.888
        & \cellsecond 0.09
        & \cellbest 180 \\

        & & 0.01
        & 32.22
        & \cellsecond 0.895
        & \cellbest 0.06
        & \cellsecond 176 \\

        & & 0.001
        & \cellsecond 32.23
        & \cellbest 0.897
        & 0.10
        & 156 \\

        & & 0.0001
        & \cellbest 32.26
        & 0.891
        & 0.10
        & 155 \\

        \cmidrule(lr){2-7}

        & \multirow{3}{*}{\centering $\epsilon_m$}
        & 0.005
        & \cellsecond 32.18
        & \cellsecond 0.889
        & 0.10
        & 156 \\

        & & 0.01
        & \cellbest 32.22
        & \cellbest 0.895
        & \cellbest 0.06
        & \cellbest 176 \\

        & & 0.1
        & 32.05
        & 0.887
        & \cellbest 0.06
        & \cellsecond 166 \\

        \midrule

        \multirow{3}{*}{LaDAR}
        & \multirow{3}{*}{\centering $\epsilon_{\mathcal{D}}$}
        & 0.0015
        & \cellbest 32.39
        & 0.891
        & 0.08
        & \cellsecond 219 \\

        & & 0.002
        & \cellsecond 32.34
        & \cellsecond 0.893
        & \cellsecond 0.07
        & \cellbest 231 \\

        & & 0.003
        & 32.28
        & \cellbest 0.896
        & \cellbest 0.06
        & 218 \\

        \bottomrule
    \end{tabularx}
\end{table*}
\subsection{Hyperparameter Sensitivity Analysis}
\label{sec:param_analysis}

We further examine the sensitivity of the key hyperparameters in AAD and LaDAR, and additionally analyze the inherent characteristics of these two modules. Quantitative results under different parameter settings are reported in Tab.~\ref{tab:ablation_key_hyperparameters}.

\Paragraph{Effect of the Sparsity Weight.} After introducing AAD, the training objective includes an additional sparsity regularization term, whose strength is controlled by the sparsity-loss weight $\lambda_s$. This term constrains the activation of dynamic anchors during optimization, thereby encouraging the representation to use anchors more selectively. 
To analyze the influence of this constraint and avoid interference from LaDAR, we disable LaDAR and vary $\lambda_s$. Imposing stronger sparsity regularization encourages the model to mask a larger number of anchors, thereby producing a sparser representation. 
As shown in Tab.~\ref{tab:ablation_key_hyperparameters}, while more aggressive anchor masking (e.g., $\lambda_s = 0.1$) can slightly improve rendering speed by reducing the number of active anchors, it also removes informative components that are necessary for accurate scene representation, leading to a noticeable degradation in reconstruction quality. By contrast, when sparsity is only weakly enforced (e.g., $\lambda_s = 0.0001$), the regularization becomes insufficient and redundant anchors remain active. Although this setting yields a slight reconstruction fidelity improvement, it also increases storage overhead and reduces rendering speed, resulting in inferior overall efficiency. Based on the above analysis, an appropriate sparsity weight is essential, as it allows AAD to retain sufficient informative anchors for accurate scene modeling, thereby improving its overall effectiveness. In particular, setting $\lambda_s = 0.01$ achieves the most favorable trade-off among reconstruction quality, storage efficiency, and rendering speed. 

\Paragraph{Effect of the Masking Threshold.} In AAD, the masking threshold $\epsilon_m$ determines whether an anchor should be suppressed during anchor activation: A larger masking threshold induces more aggressive anchor suppression, resulting in fewer active anchors, while a smaller threshold preserves denser anchor activation. To examine the sensitivity of AAD to the masking threshold, we vary $\epsilon_m$ while disabling LaDAR. Tab.~\ref{tab:ablation_key_hyperparameters} reveals clear differences in anchor reservation under varying thresholds. Among the 40,460 anchors, the variant with $\epsilon_m = 0.1$ retains 34,506 active anchors, whereas lowering the threshold to $\epsilon_m = 0.01$ and $\epsilon_m = 0.005$ increases this number to 38,879 and 39,477, respectively. Such noticeable differences in anchor quantity directly affect the overall representational quality. Specifically, when the threshold is too large (e.g., $\epsilon_m = 0.1$), AAD may discard anchors that still contain useful scene information, thereby weakening the expressive capacity of the representation and causing a clear performance drop. When the threshold is relaxed (e.g., $\epsilon_m = 0.005$), more potentially informative anchors are preserved, resulting in a more complete scene representation. However, the moderate threshold $\epsilon_m = 0.01$ provides the most favorable operating point, which achieves better reconstruction quality while maintaining lower storage cost and faster rendering speed, indicating that AAD benefits from an appropriately calibrated masking threshold rather than simply retaining more anchors.

\Paragraph{Effect of the Recalibration Threshold.} We analyze the influence of the recalibration discrepancy threshold $\epsilon_{\mathcal{D}}$ used in LaDAR. To isolate the effect of $\epsilon_{\mathcal{D}}$, AAD is disabled and only the recalibration threshold is varied. Quantitative results are reported in Tab.~\ref{tab:ablation_key_hyperparameters}, and the quality--storage trade-off is visualized in Fig.~\ref{fig:psnr_storage_curve}. The storage spikes correspond to recalibration events, each introducing about 0.03~MB overhead for the corresponding frame. As expected, a lower $\epsilon_{\mathcal{D}}$ triggers more frequent recalibration: the number of recalibration events is nearly 1{,}400 for $\epsilon_{\mathcal{D}}=0.0015$, 539 for $\epsilon_{\mathcal{D}}=0.002$, and 127 for $\epsilon_{\mathcal{D}}=0.003$.
An overly large threshold delays recalibration until substantial latent discrepancies have accumulated. For example, $\epsilon_{\mathcal{D}}=0.003$ leads to a 0.06~dB drop in PSNR compared with $\epsilon_{\mathcal{D}}=0.002$. Conversely, overly frequent recalibration brings limited quality gain but increases storage and computation. Compared with $\epsilon_{\mathcal{D}}=0.002$, setting $\epsilon_{\mathcal{D}}=0.0015$ improves PSNR by only 0.05~dB, while increasing storage overhead by about 14.3\% and reducing rendering speed.

\begin{figure}[t]
  \centering
  \includegraphics[width=\columnwidth,
  trim=0 3 0 0,
    clip
  ]{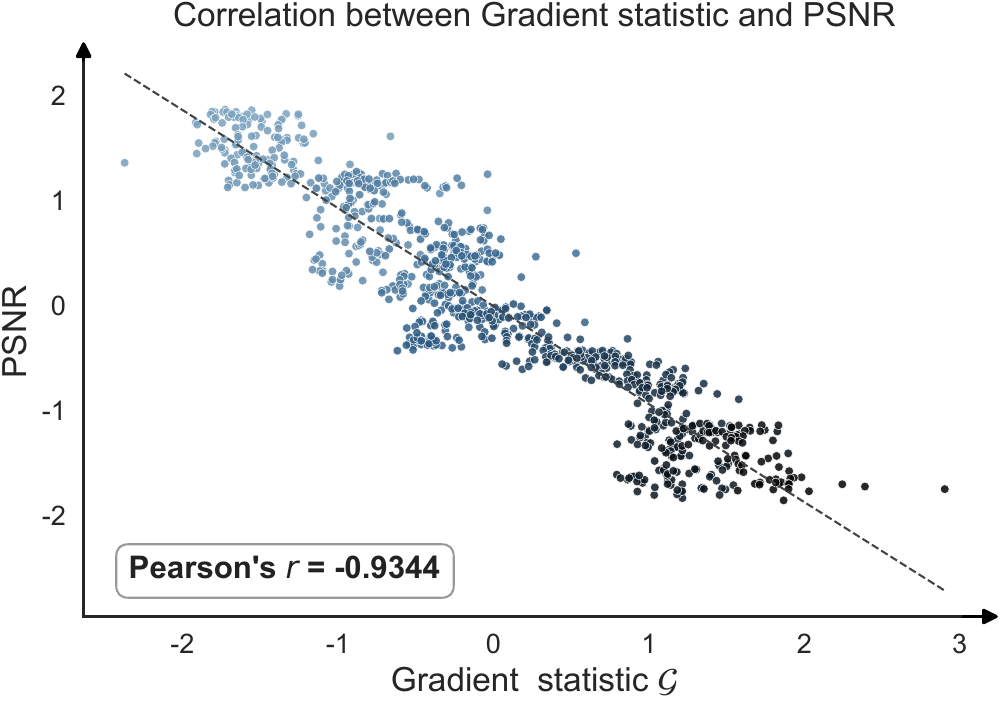}
  \caption{
    \textbf{Correlation between normalized PSNR and normalized gradient statistic across a 1,000-frame sequence on the Bar dataset.} 
   }
  \label{Fig:correlation}
\end{figure}

\subsection{Evaluation of Gradient-Based Indicator}
\label{sec:experiment:correlation}

In a streaming scenario, explicitly estimating the influence of the current frame on all subsequent frames is computationally prohibitive, since it would require either solving a frame-wise bit-allocation problem or repeatedly evaluating future reconstruction errors. Therefore, instead of directly optimizing a per-frame Lagrange multiplier $\lambda_i$ or estimating the future influence parameters, LaDAR adopts a lightweight discrepancy-driven criterion to decide when the Gaussian-attribute network $\mathbf{N}_G$ should be recalibrated.

Specifically, we use the gradient statistic $\mathcal{G}$ generated during the BTC transformation of the latent feature $f$ as a streaming-friendly indicator of the discrepancy between the current latent representation and the fixed Gaussian-attribute network $\mathbf{N}_G$. The motivation is as follows. During BTC optimization, the gradient magnitude reflects how strongly the current latent feature needs to be adjusted to fit the reconstruction objective. A larger gradient therefore indicates that the current optimization state is less compatible with the existing Gaussian-attribute decoder. In our setting, this incompatibility corresponds to a larger mismatch between the evolving latent feature $f$ and the fixed network $\mathbf{N}_G$, which may result in inaccurate Gaussian attributes and subsequently amplify reconstruction errors in later frames. Importantly, $\mathcal{G}$ is not intended to be a closed-form solution to the theoretical rate-distortion allocation problem. Instead, it serves as a practical surrogate for latent discrepancy term that is difficult to observe directly in a streaming pipeline.

Since the true latent discrepancy between the evolving feature $f$ and the fixed Gaussian-attribute network $\mathbf{N}_G$ is not directly observable, we evaluate its realized effect through the final rendering quality. The rationale is that a mismatch in the latent space affects the decoded Gaussian attributes, which further influences the reconstruction quality. Therefore, PSNR can be used as a reconstruction-level proxy for the observable consequence of latent discrepancy: a lower PSNR usually indicates that the mismatch between $f$ and $\mathbf{N}_G$ has produced a more severe degradation in the rendered result.

Based on this consideration, we select a representative 1,000-frame sequence from the \textit{Bar} dataset and record both the PSNR and the corresponding gradient statistic $\mathcal{G}$ for each frame. As shown in Fig.~\ref{Fig:correlation}, PSNR exhibits a strong negative correlation with $\mathcal{G}$, with a Pearson correlation coefficient of $-0.9344$. This result indicates that larger gradient statistics consistently correspond to lower rendering quality, which reflects greater discrepancy, and supports the use of $\mathcal{G}$ as an effective indicator of latent discrepancy in streaming setting.

We further emphasize that this correlation analysis is not used as the sole evidence for temporal importance. Rather, it validates whether the proposed gradient statistic can capture the observable effect of latent mismatch. The effectiveness of using this statistic to guide adaptive recalibration is further demonstrated by the quantitative comparisons in Sec.~\ref{sec:experiment:results} and the ablation studies in Sec.~\ref{sec:experiment:ablation}. These results jointly show that the proposed gradient-based indicator can effectively reflect the discrepancy between the evolving latent features and the Gaussian-attribute network, and can provide a reliable criterion for triggering adaptive recalibration in LaDAR.

\begin{figure}[ht]
\centering

\begin{minipage}[t]{\linewidth}
    \centering
    \includegraphics[
        width=0.92\linewidth,
        height=0.16\textheight,
        keepaspectratio,
        trim=8 6 8 6,
        clip
    ]{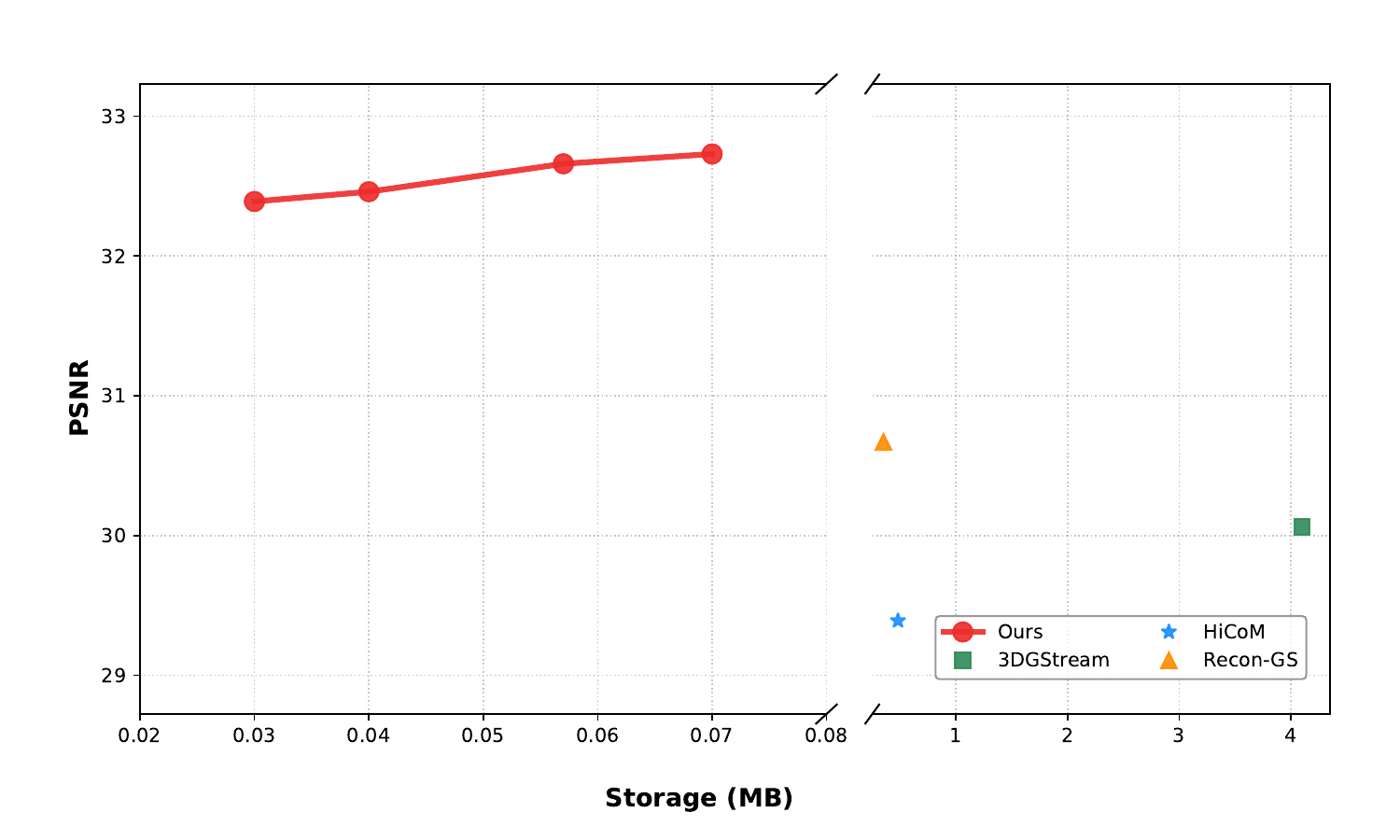}
    \vspace{1pt}
    
    {\footnotesize \textit{(a) Discussion}}
\end{minipage}

\vspace{4pt}

\begin{minipage}[t]{\linewidth}
    \centering
    \includegraphics[
        width=0.92\linewidth,
        height=0.16\textheight,
        keepaspectratio,
        trim=8 6 8 6,
        clip
    ]{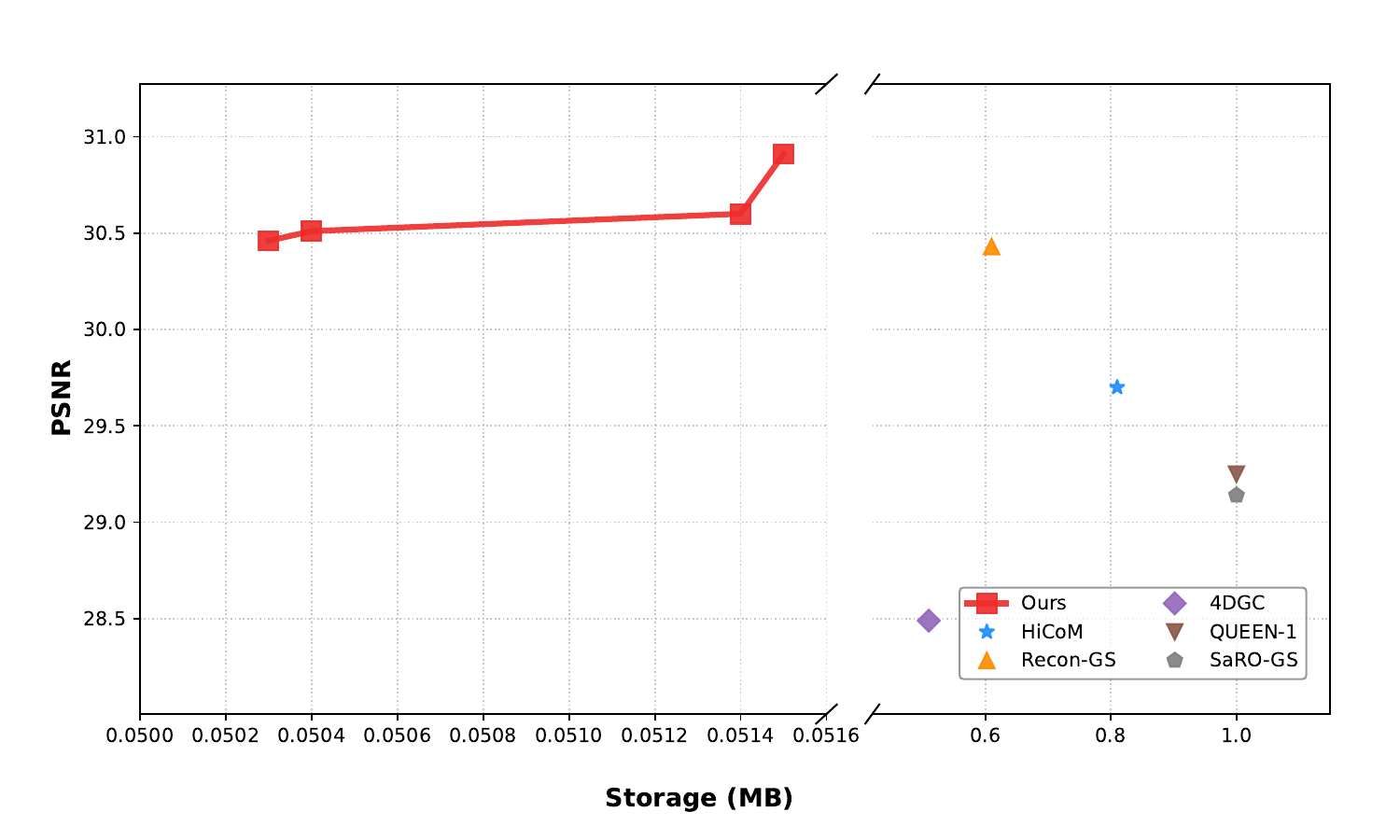}
    \vspace{1pt}
    
    {\footnotesize \textit{(b) Flame Salmon}}
\end{minipage}

\caption{\textbf{Rate-distortion performance in two scenarios.}
The proposed framework achieves a better rate-distortion trade-off than state-of-the-art baselines and provides flexible storage--quality adaptation.}
\label{fig:rd}
\end{figure}
\subsection{Rate-Distortion Performance}
\label{sec:rd_performance}

To evaluate the flexibility of \textbf{SoLAR} under different storage budgets, we instantiate a series of model variants with different storage overheads by adjusting the capacity of the transform network. This modification changes the representational budget of the model while preserving the overall framework, thereby enabling a direct examination of the rate-distortion trade-off. Fig.~\ref{fig:rd} presents the resulting rate-distortion curves and compares \textbf{SoLAR} with existing SOTA approaches~\cite{furecon,sun20243dgstream,queen,sarogs,hu20254dgc,gao2024hicom} on two scenes. A particularly notable trend appears in the extremely low-bitrate regime. Specifically, under a storage budget of only 0.03~MB, \textbf{SoLAR} still achieves a reconstruction quality of 32.39~dB. By contrast, 3DGStream requires 4.10~MB to reach 30.06~dB. This comparison indicates that \textbf{SoLAR} retains competitive reconstruction fidelity even under highly constrained storage, achieving a PSNR advantage of more than 2~dB while reducing storage by over two orders of magnitude.
More broadly, across different storage regimes, \textbf{SoLAR} consistently preserves favorable reconstruction quality, indicating that its performance degrades gracefully as the storage budget becomes tighter. Taken together, these results suggest that the proposed framework is well suited to supporting a flexible trade-off between reconstruction fidelity and resource consumption, which is desirable for deployment under heterogeneous system budgets.

\begin{figure*}[t]
  \centering
  \setlength{\tabcolsep}{0pt}

  \begin{minipage}[t]{0.24\textwidth}
    \centering
    \includegraphics[
      width=0.95\linewidth,
      height=0.14\textheight,
      keepaspectratio,
      trim=10 8 10 8,
      clip
    ]{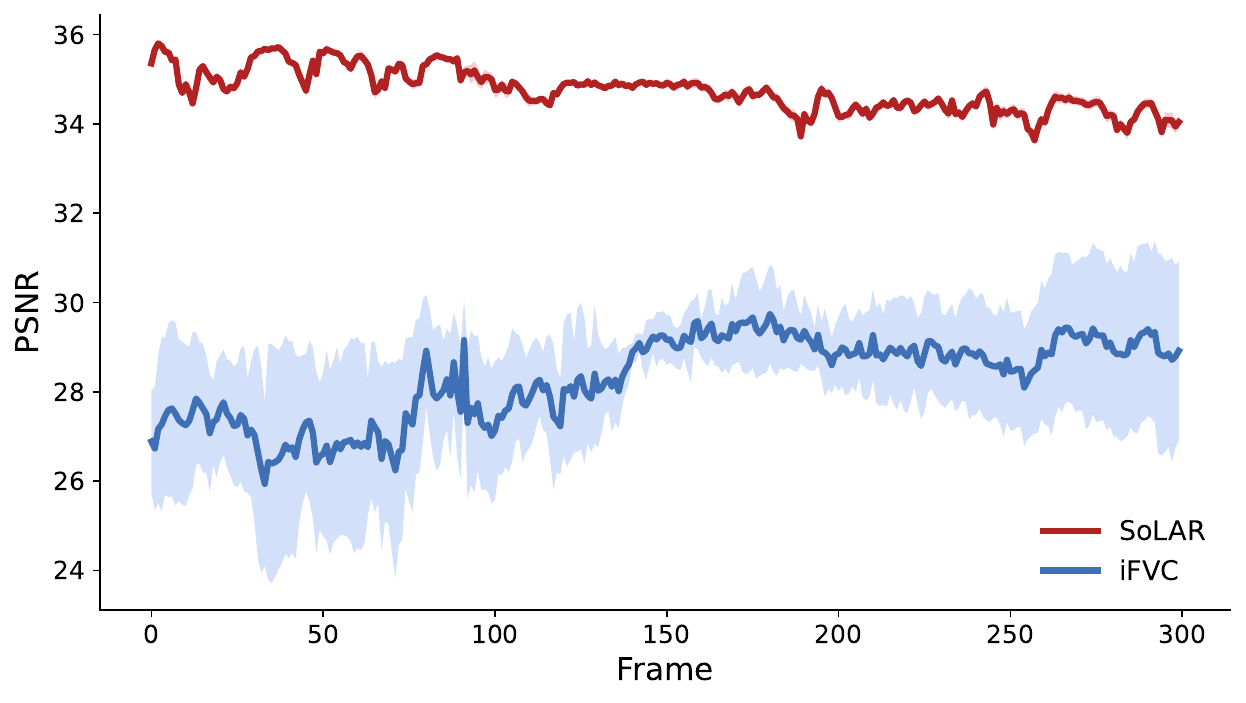}
    \vspace{1pt}
    {\footnotesize \textit{(a) Cook Spinach}}
  \end{minipage}
  \hfill
  \begin{minipage}[t]{0.24\textwidth}
    \centering
    \includegraphics[
      width=0.95\linewidth,
      height=0.14\textheight,
      keepaspectratio,
      trim=10 8 10 8,
      clip
    ]{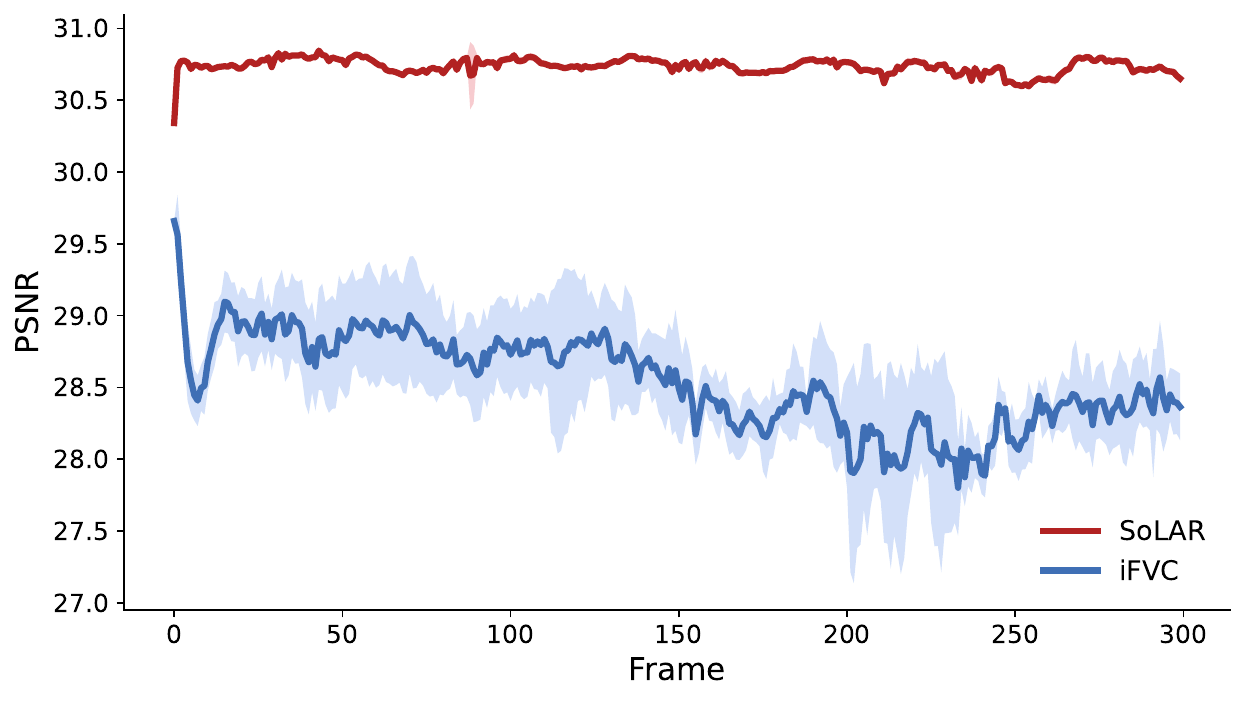}
    \vspace{1pt}
    {\footnotesize \textit{(b) Flame Salmon}}
  \end{minipage}
  \hfill
  \begin{minipage}[t]{0.24\textwidth}
    \centering
    \includegraphics[
      width=0.95\linewidth,
      height=0.14\textheight,
      keepaspectratio,
      trim=10 8 10 8,
      clip
    ]{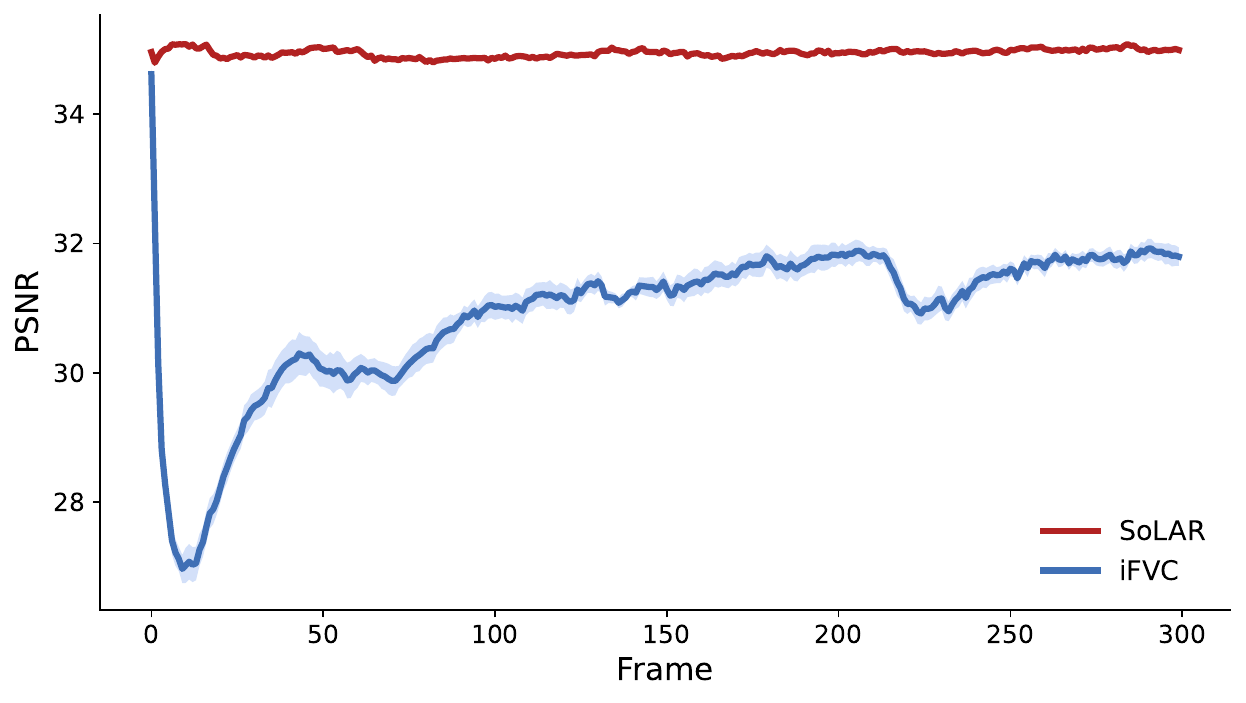}
    \vspace{1pt}
    {\footnotesize \textit{(c) Trimming}}
  \end{minipage}
  \hfill
  \begin{minipage}[t]{0.24\textwidth}
    \centering
    \includegraphics[
      width=0.95\linewidth,
      height=0.14\textheight,
      keepaspectratio,
      trim=10 8 10 8,
      clip
    ]{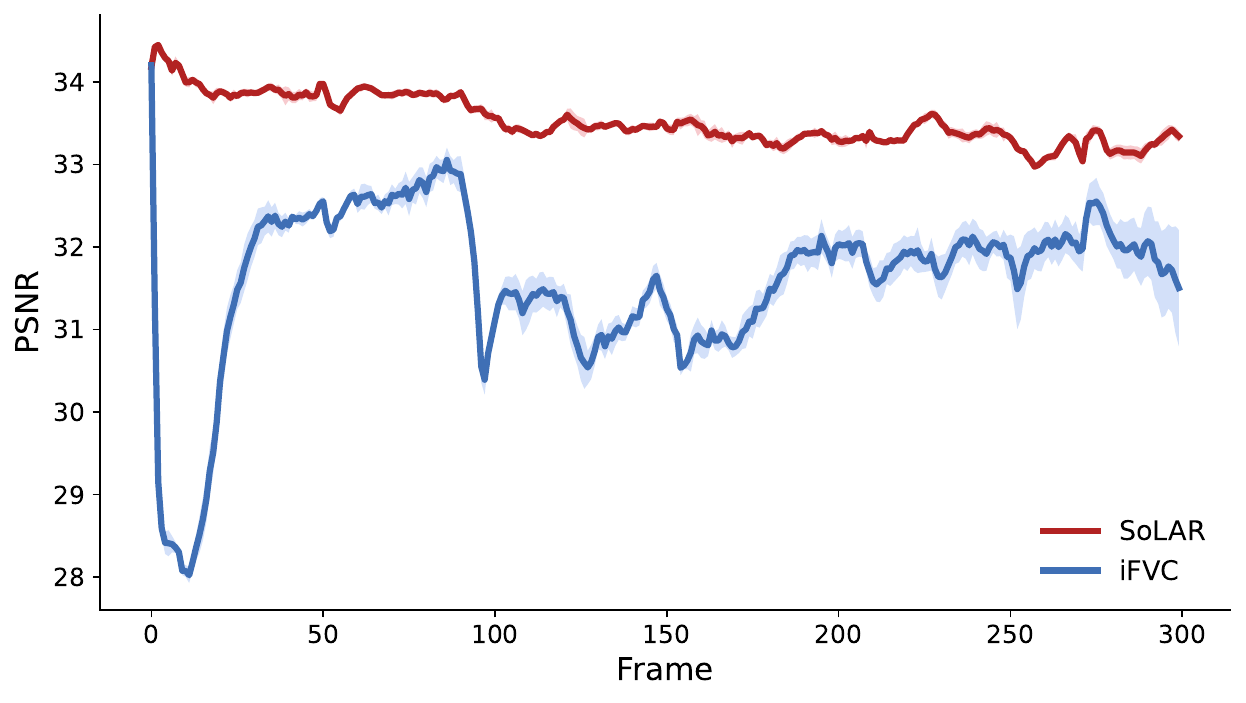}
    \vspace{1pt}
    {\footnotesize \textit{(d) Vrheadset}}
  \end{minipage}

  \vspace{-2pt}
  \caption{
    \textbf{Temporal and statistical stability over 300-frame streaming sequences.}
    Frame-wise reconstruction performance over the 300-frame sequence, aggregated over five independent runs.
    The solid line denotes the mean at each frame, and the shaded region indicates one standard deviation across runs.
    Across all scenes, \textbf{SoLAR} consistently yields higher reconstruction quality with lower variance, indicating superior robustness and stability.
  }
  \label{fig:stability}
\end{figure*}
\begin{table}[t]
\centering
\caption{\textbf{Stability comparison across repeated runs.}
}

\label{tab:stability}
\setlength{\tabcolsep}{5pt}
\renewcommand{\arraystretch}{1.15}
\begin{tabular}{c *{6}{c}}
\toprule
& \multicolumn{3}{c}{iFVC} & \multicolumn{3}{c}{SoLAR} \\
\cmidrule(lr){2-4} \cmidrule(lr){5-7}
Scene & $\mu_{\mathrm{seq}}\uparrow$ & $\overline{\sigma}_{\mathrm{run}}\downarrow$ &  $\overline{\sigma}_{\mathrm{temp}}\downarrow$
      & $\mu_{\mathrm{seq}}\uparrow$ & $\overline{\sigma}_{\mathrm{run}}\downarrow$ & $\overline{\sigma}_{\mathrm{temp}}\downarrow$ \\
\midrule
Cook Spinach & 28.29 & 1.376 & 0.932 & \textbf{34.75} & \textbf{0.085} & \textbf{0.467} \\
Flame Salmon & 28.54 & 0.312 & 0.323 & \textbf{30.74} & \textbf{0.020} & \textbf{0.052} \\
Trimming     & 30.89 & 0.182 & 1.136 & \textbf{34.94} & \textbf{0.018} & \textbf{0.060} \\
Vrheadset    & 31.60 & 0.197 & 0.971 & \textbf{33.53} & \textbf{0.051} & \textbf{0.287} \\
\bottomrule
\end{tabular}
\end{table}
\subsection{Stability Analysis}
\label{stability_analysis}

We evaluate the stability of \textbf{SoLAR} from two complementary perspectives: temporal stability and statistical stability. Temporal stability measures whether reconstruction quality can be maintained consistently along the sequence, and therefore reflects the ability of a streaming reconstruction method to resist temporal drift and error accumulation. Statistical stability, in contrast, measures the sensitivity of the method to repeated runs under identical settings. This aspect is particularly relevant for 3DGS-based methods, where stochastic optimization may still lead to non-negligible run-to-run variation even when the experimental configuration is fixed. To jointly assess these two aspects, we conduct five independent full-sequence runs for \textbf{SoLAR} and the baseline iFVC on four representative scenes. In each run, the model is evaluated over the complete 300-frame sequence. Fig.~\ref{fig:stability} reports the frame-wise reconstruction performance averaged over the five repeated runs, where the solid curve denotes the per-frame mean and the shaded band indicates the corresponding standard deviation. Tab.~\ref{tab:stability} further summarizes three complementary statistics: (1) $\mu_{\mathrm{seq}}$, the mean reconstruction performance over the 300-frame sequence, averaged across the five independent runs; (2) $\overline{\sigma}_{\mathrm{run}}$, the mean per-frame standard deviation across repeated runs, which measures optimization-induced variability; and (3) $\overline{\sigma}_{\mathrm{temp}}$, the temporal standard deviation of reconstruction performance over the full sequence, averaged across the five runs, which measures sequence-level quality fluctuation.

The results show that \textbf{SoLAR} improves not only the average reconstruction quality but also both forms of stability. Across all four scenes, \textbf{SoLAR} achieves consistently higher $\mu_{\mathrm{seq}}$ than iFVC. The improvement is particularly pronounced on \textit{Cook Spinach} and \textit{Trimming}, where the average PSNR increases from 28.29~dB to 34.75~dB and from 30.89~dB to 34.94~dB, respectively. More importantly, these quality gains are accompanied by substantially lower run-to-run variation. For example, on \textit{Cook Spinach}, $\overline{\sigma}_{\mathrm{run}}$ is reduced from 1.376~dB for iFVC to 0.085~dB for \textbf{SoLAR}; on \textit{Flame Salmon}, it decreases from 0.312~dB to 0.020~dB; and on \textit{Trimming}, it decreases from 0.182~dB to 0.018~dB. These reductions indicate that \textbf{SoLAR} is substantially less sensitive to stochastic variations during optimization and can produce more repeatable reconstruction behavior under identical settings. 
A similar advantage is observed in temporal stability. Compared with iFVC, \textbf{SoLAR} consistently yields lower $\overline{\sigma}_{\mathrm{temp}}$ across all scenes, indicating that its reconstruction quality fluctuates less over the 300-frame sequence. This trend is especially evident on \textit{Trimming}, where $\overline{\sigma}_{\mathrm{temp}}$ decreases from 1.136~dB to 0.060~dB, and on \textit{Vrheadset}, where it decreases from 0.971~dB to 0.287~dB. Since temporal fluctuation in streaming reconstruction is closely related to accumulated transformation errors, the reduced $\overline{\sigma}_{\mathrm{temp}}$ suggests that \textbf{SoLAR} more effectively suppresses error propagation and maintains a stable reconstruction trajectory throughout the sequence. Therefore, the stability results demonstrate that \textbf{SoLAR} reduces variability across repeated optimizations and preserves more consistent quality over time.

\section{Conclusion and Future Work}
\label{sec:conclusion}
In this paper, we present \textbf{SoLAR}, the first error-resilient streamable FVV framework that maintains stable reconstruction quality over long sequences without requiring GOP partitioning, thereby effectively avoiding the severe performance degradation commonly observed in existing approaches. Specifically, \textbf{SoLAR} introduces Anchor Activation Dynamics to better capture complex non-rigid transformations, which improves both scene compactness and representational capacity. Furthermore, inspired by bit allocation theory, dynamic-anchor-based volumetric video representation is analyzed within a RD optimization framework under rate control, based on which Latent Discrepancy Aware Recalibration is proposed to identify discrepancies between latent representations and recalibrate the correspondences encoded in the network. This design effectively mitigates error propagation in LFVV while preserving real-time performance and storage efficiency. 
Extensive experiments demonstrate that \textbf{SoLAR} achieves state-of-the-art reconstruction quality with minimal storage overhead, highlighting its effectiveness for stable and efficient long-horizon streamable FVV reconstruction. By addressing error propagation without relying on GOP partitioning, \textbf{SoLAR} provides a practical step toward the deployment of immersive streaming systems.

Despite these promising results, several limitations remain. First, the overall performance of \textbf{SoLAR} is sensitive to the quality of the initial I-frame, since errors in the initial representation may propagate through subsequent predictions. Second, although the proposed framework substantially improves long-term stability, performance degradation can still occur under extremely severe motion patterns, especially when abrupt scene changes exceed the modeling capacity of the current architecture. Future work will therefore focus on improving the representation quality of the initial I-frame and further enhancing the adaptive modeling capability of dynamic anchors, with the goal of achieving excellent robustness under highly dynamic long-horizon scenarios.

\bibliographystyle{IEEEtran}
\bibliography{main}

\end{document}